\documentclass[10pt,twocolumn,letterpaper]{article}

\usepackage[pagenumbers]{cvpr} 

\usepackage{kotex}                        
\usepackage[export]{adjustbox}            
\usepackage[flushleft]{threeparttable}    
\usepackage{makecell}                     
\usepackage{comment}
\usepackage{graphicx}
\usepackage{subcaption}
\usepackage{bm}
\usepackage{multirow}
\usepackage{pifont}
\usepackage[table]{xcolor}                
\usepackage{amsmath}
\usepackage{courier}
\usepackage{algorithm}
\usepackage{mathtools}
\usepackage{algpseudocode}
\usepackage{tocloft}
\usepackage[accsupp]{axessibility}
\definecolor{cvprblue}{rgb}{0.21,0.49,0.74}
\usepackage[pagebackref,breaklinks,colorlinks,allcolors=cvprblue]{hyperref}

\def\paperID{} 
\def\confName{CVPR}
\def\confYear{2026}

\title{VIRD: View-Invariant Representation through Dual-Axis Transformation for Cross-View Pose Estimation}

\author{
Juhye Park$^{1}$ Wooju Lee$^{1}$ Dasol Hong$^{1}$ Changki Sung$^{1}$ Youngwoo Seo$^{2}$ Dongwan Kang$^{2}$ Hyun Myung$^{1*}$ \\
$^{1}$Urban Robotics Lab, School of Electrical Engineering, KAIST \qquad $^{2}$Hanwha Aerospace \\
{\tt\small $^{1}$\{jhpark12,dnwn24,ds.hong,cs1032,hmyung\}@kaist.ac.kr $^{2}$\{youngwoo.seo,dongwan.kang\}@hanwha.com}
}

\addtocontents{toc}{\protect\setcounter{tocdepth}{-1}}

\begin{document}
\maketitle
\begin{abstract}
Accurate global localization is critical for autonomous driving and robotics, but GNSS-based approaches often degrade due to occlusion and multipath effects.
As an emerging alternative, cross-view pose estimation predicts the 3-DoF camera pose corresponding to a ground-view image with respect to a geo-referenced satellite image. However, existing methods struggle to bridge the significant viewpoint gap between the ground and satellite views mainly due to limited spatial correspondences. We propose a novel cross-view pose estimation method that constructs view-invariant representations through dual-axis transformation (VIRD). VIRD first applies a polar transformation to the satellite view to facilitate horizontal correspondence, then uses context-enhanced positional attention on the ground and polar-transformed satellite features to mitigate vertical misalignment, explicitly bridging the viewpoint gap. To further strengthen view invariance, we introduce a view-reconstruction loss that encourages the derived representations to reconstruct the original and cross-view images. Experiments on the KITTI and VIGOR datasets demonstrate that VIRD outperforms the state-of-the-art methods without orientation priors, reducing median position and orientation errors by 50.7\% and 76.5\% on KITTI, and 18.0\% and 46.8\% on VIGOR, respectively.
\end{abstract}
\section{Introduction}
\label{sec:intro}

Global localization plays a critical role in autonomous driving and mobile robotics, enabling systems to accurately navigate in real-world environments~\cite{thrun2002probabilistic, lee20232, yin2024survey, yang2021cross, shi2019spatial}. Although the global navigation satellite system (GNSS) is widely adopted for outdoor localization~\cite{lemmens2011global, congram2021relatively}, its reliability degrades in dense urban areas due to signal occlusion and multipath effects~\cite{ben2011improving, xia2020geographically, xia2021cross}. 

\begin{figure}[ht]
    \centering
    \includegraphics[width=\columnwidth]{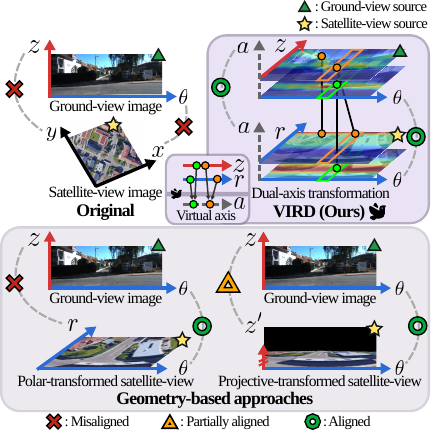}
    \caption{
    Ground-view and satellite-view images exhibit a large viewpoint gap due to misalignment along both the horizontal and vertical axes on the image plane. Previous geometry-based methods, such as polar and projective transformations, partially mitigated this issue by ensuring horizontal correspondence but could not fully address vertical misalignment. For example, projective transformations often produce severe artifacts around vertical structures such as buildings. VIRD overcomes these limitations through dual-axis transformation, introducing a shared virtual vertical axis $a$ to establish consistent cross-view correspondence.
    }
    \label{fig:figure1}
\end{figure}

Cross-view pose estimation (CVPE) has emerged as a promising alternative to GNSS-based localization. It estimates the fine-grained 3 degrees of freedom (DoF) camera pose corresponding to a ground-view image by leveraging a geo-referenced satellite image~\cite{shi2022beyond, shi2023boosting, wang2023fine, lee2025pidloc, lentsch2023slicematch, xia2023convolutional, song2023learning, xia2025fg}. Early studies often assume coarse orientation priors and refine the pose through iterative optimization within a narrow search space~\cite{shi2022beyond, shi2023boosting, wang2023fine, lee2025pidloc}. 
However, such priors are often inaccurate or unavailable in practice, causing these methods to converge to suboptimal solutions and to suffer from degraded performance.
Thus, recent methods have explored omnidirectional CVPE, which searches the full 360-degree orientation space without requiring orientation priors~\cite{sarlin2023orienternet, lentsch2023slicematch, xia2023convolutional}. To support exhaustive search, these approaches typically extract orientation-aware descriptors from the ground image and match them with satellite descriptors sampled across candidate poses, which improves robustness to local minima~\cite{lentsch2023slicematch, xia2023convolutional}.

However, existing omnidirectional CVPE methods often neglect the significant viewpoint gap between ground and satellite views, a fundamental challenge for accurate cross-view matching. To address this issue, some methods have adopted content-based cross-attention~\cite{lentsch2023slicematch} or contrastive learning~\cite{lentsch2023slicematch, xia2023convolutional}. However, these approaches focus primarily on semantic similarity and often ignore spatial correspondences, leaving the viewpoint gap largely unresolved. Geometry-based approaches have also been explored to address this gap through polar~\cite{shi2019spatial, shi2020looking} and projective transformations~\cite{shi2022beyond, shi2023boosting}.
Polar transformations transform the satellite view into ground-aligned horizontal coordinates, but neglect vertical-axis misalignment~\cite{shi2019spatial,shi2020looking}.
Projective transformations, parameterized by camera geometry, aim to simultaneously rectify both horizontal and vertical axes. However, they are highly sensitive to camera calibration errors and prone to severe projection artifacts around vertical structures (\eg, buildings) when depth information is unreliable~\cite{shi2022beyond, shi2023boosting}. Consequently, effectively resolving vertical misalignment remains an open challenge in these geometry-based methods, as qualitatively shown in \cref{fig:figure1}.

To overcome these limitations, we propose VIRD (\textbf{V}iew-\textbf{I}nvariant \textbf{R}epresentation through \textbf{D}ual-axis transformation), a novel omnidirectional CVPE method that effectively bridges the viewpoint gap between ground and satellite views.
VIRD constructs view-invariant descriptors by explicitly transforming cross-view features along horizontal and vertical axes. For the horizontal axis, we apply a polar transformation to the satellite view, following prior works~\cite{shi2019spatial,shi2020looking}. To resolve the remaining vertical misalignment, we propose context-enhanced positional attention (CEPA), which transforms the ground and polar-transformed satellite features along the vertical axis via positional attention mechanisms~\cite{de2024positional}. CEPA learns a view-consistent vertical transformation without camera parameters and adaptively captures vertical structures in the ground view by leveraging contextual cues. The view-invariant descriptors are further refined by a view-reconstruction loss that guides the model to reconstruct the original and cross-view images during training.

Our contributions can be summarized as follows: 
\begin{itemize}
    \item A novel omnidirectional CVPE framework constructs view-invariant descriptors to overcome the significant viewpoint gap and enable accurate cross-view matching.
    \item A dual-axis transformation strategy addresses horizontal and vertical misalignment using polar transformation and context-enhanced positional attention, reducing reliance on camera parameters.
    \item A view-reconstruction loss enhances the view invariance of descriptors by training the model to reconstruct the original and cross-view images.
    \item The proposed method achieved state-of-the-art performance on the KITTI and VIGOR datasets without orientation priors, reducing median position and orientation errors by 50.7\% and 76.5\% on KITTI, and 18.0\% and 46.8\% on VIGOR, respectively.
\end{itemize}
\newcommand{\Pose}[1]{\mathbf{p}_{#1}}
\newcommand{\AttnW}[1]{\mathcal{A}_{#1}}

\begin{figure*}[ht]
    \centering
    \includegraphics[width=\textwidth]{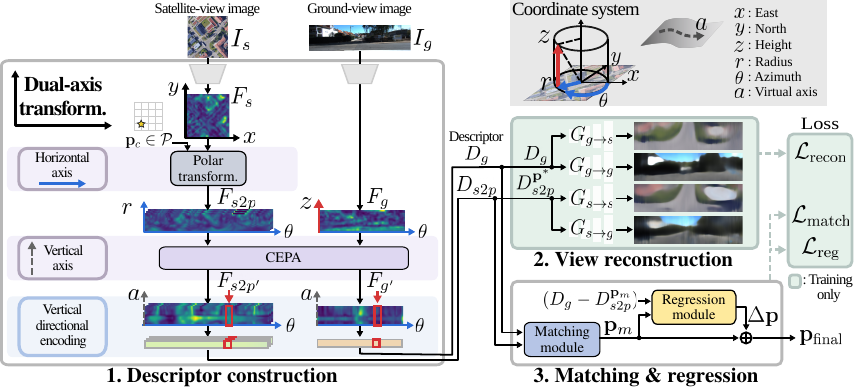}
    \caption{
    Overview of VIRD. VIRD constructs view-invariant descriptors through dual-axis transformation. The process begins by applying a polar transformation to transform the satellite features $F_s$ into a horizontally corresponding representation to the ground features $F_g$ for each candidate pose $\Pose{c} \in \mathcal{P}$ (the set of candidate poses), generating the polar-transformed features $F_{s2p}$. The context-enhanced positional attention (CEPA) module then vertically transforms cross-view features through a positional attention mechanism. The resulting satellite and ground features, $F_{s2p'}$ and $F_{g'}$, are compressed along the vertical axis to generate orientation-aware descriptors $D_{s2p}$ and $D_g$. 
    During training, a view-reconstruction loss is computed at the ground-truth pose $\Pose{}^*$ to enforce view invariance by reconstructing both original and cross views. At inference, the final 3-DoF pose is estimated by combining descriptor matching with residual regression.
    }
    \label{fig:figure2}
\end{figure*}

\section{Related Works}
\label{sec:related}

\paragraph{Limited-angle cross-view pose estimation.}
Limited-angle cross-view pose estimation estimates the 3-DoF camera pose corresponding to a ground-view image within a geo-referenced satellite image, given a coarse orientation prior. 
This task was first addressed by HighlyAccurate~\cite{shi2022beyond}, which refined the pose iteratively using the Levenberg–Marquardt (LM) algorithm~\cite{levenberg1944method, marquardt1963algorithm}. This optimization-based paradigm was later extended using either LM~\cite{shi2022beyond, wang2023satellite, wang2023view, wang2024view} or neural pose optimizers~\cite{shi2023boosting, wang2023fine, lee2025pidloc}.
However, these methods remain constrained by their reliance on orientation priors, limiting the search space to a narrow angular range.

\paragraph{Omnidirectional cross-view pose estimation.}
Omnidirectional cross-view pose estimation eliminates the dependence on orientation priors by estimating the 3-DoF pose across the full 360-degree orientation range~\cite{xia2022visual, sarlin2023orienternet, lentsch2023slicematch, xia2023convolutional}. It is often achieved by matching a ground descriptor with satellite descriptors sampled from candidate poses. In this setting, the descriptors are required to be orientation-aware for accurate orientation estimation and view-invariant to bridge the viewpoint gap. Existing methods often address orientation awareness by leveraging the horizontal azimuthal structure of ground images, while mitigating the viewpoint gap through semantics-based approaches. For example, SliceMatch~\cite{lentsch2023slicematch} employs content-based cross-attention, while CCVPE~\cite{xia2023convolutional} adopts contrastive learning to enhance semantic consistency across views. However, these approaches focus primarily on semantic similarity and fail to explicitly model spatial correspondence between ground and satellite images, which is critical for robust pose estimation. 
To address this limitation, we propose a novel framework that explicitly models spatial correspondences between ground and satellite views, and incorporates orientation awareness into both ground and satellite descriptors, enabling robust omnidirectional pose estimation.

\paragraph{Geometry-based transformations.}
Geometry-based transformations have been widely adopted in cross-view pose estimation to reduce the viewpoint gap. Many methods apply polar transformations~\cite{shi2019spatial, shi2020looking} or projective transformations based on camera parameters~\cite{shi2022beyond, shi2023boosting}. Polar transformations transform the satellite view into a ground-aligned horizontal coordinate but leave vertical misalignment unresolved. In contrast, projective transformations suffer from projection artifacts around vertical structures such as buildings when depth is unavailable. Although recent works incorporate LiDAR data to mitigate these issues~\cite{wang2023satellite, lee2025pidloc}, such approaches are costly. 
We therefore propose a novel dual-axis transformation strategy that integrates polar transformation with context-enhanced positional attention, addressing horizontal and vertical misalignment without additional sensors.

\section{Proposed Method}
\label{sec:method}

\paragraph{Problem definition.}
Given a ground-view image $I_g$ and a satellite-view image $I_s$, the goal is to estimate the 3-DoF pose $\Pose{} = (x, y, \theta)$ of the ground camera. $(x, y)\in \mathbb{R}^2$ denotes the camera's position within $I_s$, where $x$ and $y$ are aligned with east and north, respectively. $\theta\in[-\pi, \pi)$ is the yaw angle measured clockwise from east. The ground image may be either panoramic or a limited horizontal field of view (HFoV), and is assumed to follow a cylindrical projection as in~\cite{xia2023convolutional}. Following previous works~\cite{lentsch2023slicematch, xia2023convolutional}, we assume that the camera's pitch and roll are negligible.

\paragraph{Overview.} 
The proposed CVPE method, \textbf{VIRD}, is designed to bridge the viewpoint gap by constructing view-invariant descriptors.
As illustrated in \cref{fig:figure2}, this framework consists of three components: (1) the construction of view-invariant descriptors through dual-axis transformation, followed by vertical directional encoding (\cref{sec:oavi}); (2) a view-reconstruction loss that enhances the view invariance of descriptors by reconstructing both original and cross-view images (\cref{sec:vi}); and (3) matching and regression modules that predict the final pose (\cref{sec:matching_and_regression}). 
All components are jointly optimized in an end-to-end manner under a unified multi-objective loss function (\cref{sec:training}).

\subsection{Descriptor construction}
\label{sec:oavi}

\paragraph{Feature extraction.}
Let $F_g \in \mathbb{R}^{C \times H \times W_g}$ and $F_s \in \mathbb{R}^{C \times A \times A}$ represent features extracted from the ground image $I_g$ and the satellite image $I_s$ using a pretrained convolutional backbone, respectively. Here, $C$, $H$, $W_g$, and $A$ denote the number of channels, the height and width of the ground feature map, and the spatial size of the satellite feature map, respectively. For ground images under cylindrical projection, the horizontal axis of $F_g$ represents azimuthal angles, and the vertical axis corresponds to height. For satellite images, the horizontal axis of $F_s$ represents east and the vertical axis corresponds to north, as illustrated in \cref{fig:figure2}. This inherent misalignment between the coordinate systems of $F_g$ and $F_s$ hinders cross-view matching and full 360-degree orientation search.

\begin{figure}[t]
    \centering
    \includegraphics[width=\columnwidth]{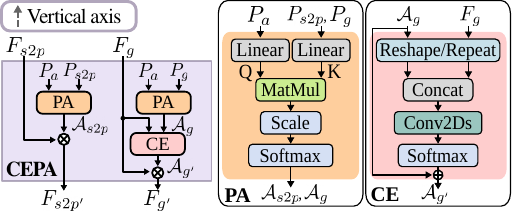}
    \caption{Schematic illustration of the context-enhanced positional attention (CEPA) module consisting of positional attention (PA) and context enhancement (CE). PA generates attention weights $\mathcal{A}_g$ and $\mathcal{A}_{s2p}$ by comparing the shared virtual positional encodings $P_a$ with the positional encodings of the ground and satellite views, $P_g$ and $P_{s2p}$, respectively. CE then adaptively refines the ground attention weights $\mathcal{A}_g$ using contextual information from the ground feature $F_g$, resulting in $\mathcal{A}_{g'}$. These attention weights vertically transform the cross views by weighting the features $F_g$ and $F_{s2p}$ with $\mathcal{A}_{g'}$ and $\mathcal{A}_{s2p}$, respectively. 
    }
    \label{fig:figure3}
\end{figure}

\paragraph{Polar transformation.}
To establish horizontal correspondence between the ground and satellite views, we apply a polar transformation to the satellite feature map $F_s$, centered at the given candidate position in pixel coordinates $(u_c^s, v_c^s)$, as in~\cite{shi2019spatial, shi2020looking}. The transformed feature $F_{s2p}^{(u_c^s, v_c^s)}\in \mathbb{R}^{C \times H \times W_s}$ maps the azimuth and radial distance to the horizontal and vertical axes, respectively, as defined below:
\begin{equation} 
\begin{aligned} 
\begin{bmatrix} u^s \\ v^s \end{bmatrix} 
= \begin{bmatrix} u_c^s \\ v_c^s \end{bmatrix}
- \rho(v^{s2p}) 
\begin{bmatrix} \cos\left(\frac{2\pi}{W_s}u^{s2p}\right) \\ \sin\left(\frac{2\pi}{W_s}u^{s2p}\right) \end{bmatrix}, 
\end{aligned} 
\end{equation}
\begin{equation}
\begin{aligned}
\rho(v) &\coloneqq \bigl(r_{\max}-r_{\min}\bigr)\!\left(1-\frac{v}{H}\right)+r_{\min},
\end{aligned}
\end{equation}
where $(u^s, v^s)$ and $(u^{s2p}, v^{s2p})$ represent the pixel coordinates of the satellite and transformed features, respectively. $r_{\min}$ and $r_{\max}$ denote the minimum and maximum radii of the polar sampling range in pixels. The transformed feature width is $W_s = \frac{2\pi}{\text{HFoV}} \cdot W_g$, ensuring consistency across varying FoVs. This transformation maps the azimuth direction as the horizontal axis of the image plane, assigning an explicit orientation axis to the satellite view and mitigating horizontal misalignment between the cross views. 

\begin{figure}[t]
    \centering
    \includegraphics[width=\columnwidth]{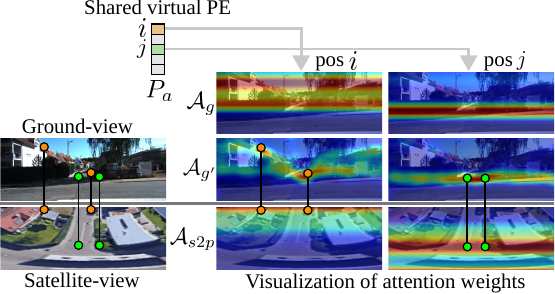}
    \caption{Visualization of attention weights. It shows the activations corresponding to each shared virtual positional encoding (PE). Red indicates stronger activations. Orange and green dots indicate the real spatial correspondence between the views.
    }
    \label{fig:figure4}
\end{figure}

\paragraph{Positional attention.}
Although the polar transformation establishes azimuthal correspondence between $F_{s2p}$ and $F_g$, their vertical axes remain unaligned. As illustrated in \cref{fig:figure2}, the vertical axis of $F_{s2p}$ corresponds to radial distance, while the vertical axis of $F_g$ represents height. Prior methods~\cite{shi2022beyond, shi2023boosting} address this with projective transformations, but they rely on camera parameters and struggle to handle vertical structures when depth is unavailable. To overcome this, we propose a positional attention mechanism that explicitly learns vertical correspondence within a shared coordinate system, eliminating the dependence on camera parameters, as shown in \cref{fig:figure3}.

Positional attention computes attention weights from fixed positional encodings and is primarily adopted for its computational efficiency~\cite{schmidt2023bridging, de2024positional}.
Here, we reinterpret it as learning a transformation between the vertical coordinates of the ground and satellite views via a shared virtual axis.
This axis is defined as a learned, view-invariant vertical coordinate system onto which ground and satellite features are projected.
Specifically, we define three positional encodings: shared virtual positional encoding $P_a \in \mathbb{R}^{H_Q \times d_p}$, ground positional encoding $P_g \in \mathbb{R}^{H_K \times d_p}$, and satellite positional encoding $P_{s2p} \in \mathbb{R}^{H_K \times d_p}$.
All encodings are sinusoidal encodings of dimension $d_p$, representing positional information of each vertical image coordinate. 
$H_Q$ represents the height of the features in the shared coordinate, and $H_K = H$ refers to the height of the features in each view.
The positional attention weights $\AttnW{v} \in \mathbb{R}^{H_Q \times H_K}$ for each view $v \in \{g, s2p\}$ are computed as:
\begin{equation}
\AttnW{v} = \text{softmax} \left( \frac{(P_aW^Q_v)(P_vW^K_v )^{\top}}{\sqrt{d_k}} \right),
\label{eq:positional_attention_weights}
\end{equation} 
where $W^Q_v$ and $W^K_v$ are the query and key weight matrices for each view, and $d_k$ is the dimension of the queries and keys. These attention weights activate the regions in each view corresponding to the same vertical position in the shared virtual positional space, enabling consistent vertical transformation across horizontal directions. As shown in \cref{fig:figure4} (first and third rows), high activations (red) consistently appear along identical road regions and extend to the vertical structures, roughly following the height of building roofs. This learning-based approach allows the model to capture the floor and the overall vertical layout of the ground view, thereby improving localization accuracy over traditional projective methods.

\paragraph{Context-enhanced positional attention.} Although positional attention partially mitigates vertical misalignment based on positional encodings, a key limitation remains: it fundamentally assumes a uniform vertical transformation across all horizontal directions, independent of the input ground context. This assumption facilitates stable learning of cross-view mappings but limits adaptability to vertical structures that vary with horizontal direction and scene context in the ground view. To address this issue, we enhance the ground attention weights $\AttnW{g}$ with local ground context. The context-enhanced ground attention weights $\AttnW{g'}\in \mathbb{R}^{H_Q\times H_K \times W_g}$ are computed as:
\begin{equation}
\AttnW{g'} = \AttnW{g} + \text{Softmax} \left( \Phi(\AttnW{g} \oplus F_g) \right),
\label{eq:context_enhancement}
\end{equation}
where $\oplus$ denotes channel-wise concatenation after shape alignment. The concatenated tensor is processed by convolutional layers $\Phi$, followed by a softmax operation along the $H_K$ dimension. 
A skip connection facilitates smooth gradient flow.
This adjustment enables the model to adaptively transform ground features based on their context, improving vertical correspondence across horizontal directions, as illustrated in \cref{fig:figure4} (second row). The pseudocode for this process is provided in the supplementary material.

These attention weights are applied to the features of each view, resulting in vertically transformed features $F_{g'}\in\mathbb{R}^{C \times H_Q \times W_g}$ and $F_{s2p'}\in\mathbb{R}^{C \times H_Q \times W_s}$ as follows: 
\begin{equation}
    \begin{aligned}
    F_{g'}[c, h_q, w] &= \sum\nolimits^{H_K}_{h_k=1} \AttnW{g'}[h_q, h_k, w] \odot F_g[c, h_k, w], \\
    F_{s2p'}[c, h_q, w] &= \sum\nolimits^{H_K}_{h_k=1} \AttnW{s2p}[h_q, h_k] \odot F_{s2p}[c, h_k, w],
    \end{aligned}
\label{eq:apply_attention_weights}
\end{equation}
where $\odot$ denotes element-wise multiplication.

\paragraph{Vertical directional encoding.} The 3D features transformed along the horizontal and vertical axes, $F_{g'}$ and $F_{s2p'}$, are compressed along the vertical direction using shared multi-layer perceptrons (MLPs) applied to each column. The resulting features are then flattened across the channel and width dimensions, while preserving the width information, following~\cite{xia2023convolutional}. 
This process yields orientation-aware 1D descriptors, $D_g \in \mathbb{R}^{K_g}$ and $D_{s2p} \in \mathbb{R}^{K_s}$, with $K_s = \frac{2\pi}{\mathrm{HFoV}} \cdot K_g$.

\begin{table*}[ht!]  
    \caption{Position and orientation estimation errors on the KITTI~\cite{geiger2013vision} dataset without orientation priors. Lon. and Med. are abbreviations for longitudinal and median, respectively. $^*$ denotes results reproduced from the official implementations. Best in \textbf{bold}, second best \underline{underlined}.} 
    \label{tab:kitti}
    \begin{centering}
    \begin{threeparttable}
    \resizebox{0.95\textwidth}{!}{
        \footnotesize{
        \begin{tabular}{lcccccccccccccc}
            \toprule
            \multirow{3}{*}{}
                & \multirow{3}{*}{}
                &
                & \multicolumn{6}{c}{Position}
                & \multicolumn{4}{c}{Orientation} \\
            \cmidrule(lr){4-9} \cmidrule(lr){10-13}
                & 
                &
                & \multicolumn{2}{c}{Error (m) ↓} 
                & \multicolumn{2}{c}{Lateral recall (\%) ↑}
                & \multicolumn{2}{c}{Lon. recall (\%) ↑}
                & \multicolumn{2}{c}{Error ($^{\circ}$) ↓} 
                & \multicolumn{2}{c}{Recall (\%) ↑} \\
                Method
                & Backbone
                & Area
                & Mean & Med. 
                & R@1m & R@5m
                & R@1m & R@5m
                & Mean & Med. 
                & R@1$^{\circ}$ & R@5$^{\circ}$ \\
            \midrule
            HighlyAccurate~\cite{shi2022beyond} 
                & VGG16
                & Same
                & 15.51 & 15.97 
                & 5.17 & 25.44
                & 4.66 & 25.39
                & 89.91 & 90.75 
                & 0.61 & 2.89 \\
            SliceMatch~\cite{lentsch2023slicematch} 
                & VGG16
                & Same
                & 9.39 & 5.41  
                & 39.73 & 87.92  
                & 13.63 & 49.22   
                & 8.71 & 4.42 
                & 11.35 & 55.82 \\
            CCVPE~\cite{xia2023convolutional} 
                & EfficientNet-B0
                & Same
                & 6.88 & 3.47  
                & 53.30 & 85.13  
                & 25.84 & 68.49   
                & 15.01 & 6.12 
                & 8.96 & 42.75 \\
            DenseFlow$^*$~\cite{song2023learning} 
                & ResNet18~\cite{he2016deep}
                & Same
                & 6.47 & 4.26  
                & 73.87 & 97.27 
                & 14.82 & 58.50   
                & \textbf{3.72} & \underline{0.99} 
                & \underline{50.62} & 90.06 \\
            FG2$^*$~\cite{xia2025fg} 
                & DINOv2~\cite{oquab2023dinov2}
                & Same
                & 5.81 & 4.26  
                & 24.97 & 75.78  
                & 24.57 & \underline{75.14}   
                & 89.64 & 90.26 
                & 0.42 & 2.73 \\
            \rowcolor{gray!20} \textbf{VIRD (Ours)} 
            & VGG16 & Same
                & \textbf{5.47} & \textbf{2.07}
                & \underline{79.46} & \underline{98.30}
                & \textbf{31.65} & \textbf{75.78}
                & 5.16 & 1.02 
                & 49.32 & \textbf{92.95} \\
            \rowcolor{gray!20} \textbf{VIRD (Ours)} 
            & EfficientNet-B0 & Same
                & \underline{5.55} & \underline{2.22}
                & \textbf{79.43} & \textbf{98.41}
                & \underline{29.84} & 73.89
                & \underline{4.87} & \textbf{0.98} 
                & \textbf{51.2} & \underline{92.82} \\
            \midrule
            HighlyAccurate~\cite{shi2022beyond} 
            & VGG16
            & Cross
                & 15.50 & 16.02 
                & 5.60 & 25.60
                & 5.64 & 25.76
                & 89.84 & 89.85 
                & 0.60 & 2.65 \\
            SliceMatch~\cite{lentsch2023slicematch} 
            & VGG16
            & Cross
                & 14.85 & 11.85  
                & 24.00 & 72.89  
                & 7.17 & 33.12   
                & \underline{23.64} & 7.96 
                & \textbf{31.69} & 31.69 \\
            CCVPE~\cite{xia2023convolutional} 
            & EfficientNet-B0
            & Cross
                & 13.94 & 10.98  
                & 23.42 & 60.46  
                & 11.81 & 42.12   
                & 77.84 & 63.84 
                & 3.14 & 14.56 \\
            DenseFlow$^*$~\cite{song2023learning} 
            & ResNet18
            & Cross
                & 25.85 & 18.84  
                & 19.80 & 47.15
                & 4.35 & 23.75  
                & 69.51 & 42.04 
                & 10.87 & 29.57 \\
            FG2$^*$~\cite{xia2025fg} 
            & DINOv2
            & Cross
                & 13.58 & 11.72  
                & 9.92 & 42.48
                & 10.95 & 43.03  
                & 90.12 & 90.42 
                & 0.65 & 2.93 \\
            \rowcolor{gray!20} \textbf{VIRD (Ours)} 
            & VGG16
            & Cross
                & \underline{12.30} & \underline{7.05}
                & \underline{43.61} & \underline{81.86}
                & \underline{12.88} & \underline{46.83}
                & 25.10 & \underline{2.22} 
                & 27.65 & \underline{69.86} \\
            \rowcolor{gray!20} \textbf{VIRD (Ours)} 
            & EfficientNet-B0
            & Cross
                & \textbf{11.12} & \textbf{5.41}
                & \textbf{45.88} & \textbf{85.49}
                & \textbf{14.97} & \textbf{52.55}
                & \textbf{22.03} & \textbf{1.87} 
                & \underline{30.39} & \textbf{74.87} \\
            \bottomrule
        \end{tabular}
        }}
    \end{threeparttable}
    \end{centering}
\end{table*}

\subsection{View-reconstruction}
\label{sec:vi}

To enhance view invariance, we introduce a view-reconstruction loss that trains each descriptor to reconstruct both the original and cross views. 
This loss encourages the model to learn a view-invariant representation that captures shared structures
between cross views.

\paragraph{Reconstruction mechanism.}
Given the ground-truth pose $\Pose{}^* = (x^*, y^*, \theta^*)$, the satellite descriptor $D_{s2p}^{(x^*, y^*)}$ obtained in \cref{sec:oavi} is horizontally shifted by $\theta^*$ and cropped to match the ground descriptor, yielding $D_{s2p}^{\Pose{}^*} \in \mathbb{R}^{K_g}$.
The aligned descriptors are reshaped and vertically expanded via MLP layers, then processed by 2D convolutional and upsampling layers to reconstruct the target view.
Each decoder $G_{i\to j}$ generates a full-resolution image of the target view $j$ from the descriptor of the source view $i$.

\paragraph{Reconstruction loss.}
The view-reconstruction loss is defined using the $\ell_1\text{-loss}$ as follows:
\begin{equation}
\begin{aligned}
\mathcal{L}_{\text{origin}}
&= \|I_g - G_{g \to g}(D_g)\|_1 + \|I_{s2p}^{\Pose{}^*} - G_{s \to s}(D_{s2p}^{\Pose{}^*})\|_1,
\\[2pt]
\mathcal{L}_{\text{cross}}
&= \|I_g - G_{s \to g}(D_{s2p}^{\Pose{}^*})\|_1 + \|I_{s2p}^{\Pose{}^*} - G_{g \to s}(D_g)\|_1,
\\[2pt]
\mathcal{L}_{\text{recon}}
&= \alpha_1 \mathcal{L}_{\text{origin}}
 + \alpha_2 \mathcal{L}_{\text{cross}},
\end{aligned}
\end{equation}
where $\Pose{}^*$ is the ground-truth pose, and $\alpha_1$, $\alpha_2$ are weighting coefficients. 
The polar-transformed satellite image $I_{s2p}^{\Pose{}^*}$ is generated following the same alignment procedure as that of $D_{s2p}^{\Pose{}^*}$. 
This loss guides the descriptors to better encode vertical structures, where the viewpoint gap is particularly large, and thus helps resolve ambiguities in visually similar road scenes.

\subsection{Matching and regression}
\label{sec:matching_and_regression}

\paragraph{Descriptor matching.}
To estimate the coarse pose, the ground descriptor is matched with satellite descriptors from candidate poses $\Pose{c} = (x_c, y_c, \theta_c)$, sampled from a predefined spatial grid. 
The similarity score for each candidate pose, $S^{\Pose{c}}$, is computed using the cosine similarity between $D_g$ and $D_{s2p}^{\Pose{c}}$.
To train discriminative descriptors, we adopt the InfoNCE loss~\cite{oord2018representation}, which encourages high similarity at the ground-truth pose $\Pose{}^*$ and penalizes other non-matching poses. During inference, the candidate pose that maximizes $S^{\Pose{c}}$ is selected as the coarse matching pose $\Pose{m}$.

\paragraph{Pose regression.}
The regression module refines the matching pose $\Pose{m}$ by predicting residuals $\Delta \Pose{} = (\Delta x, \Delta y, \Delta \theta)$ to compensate for the resolution limits of descriptor matching. The residuals are predicted from the difference between $D_g$ and $D_{s2p}^{\Pose{m}}$, together with $\Pose{m}$, and are added to produce the final pose $\Pose{\text{final}} = \Pose{m} + \Delta \Pose{}$. It is trained using the $\ell_2\text{-loss}$ computed over the pose residuals.

\subsection{Training details}
\label{sec:training}

\paragraph{Total loss.}
The final objective combines the view-reconstruction, matching, and regression losses as $\mathcal{L} = \mathcal{L}_{\text{recon}} + \mathcal{L}_{\text{match}} + \mathcal{L}_{\text{reg}}$. 
Detailed formulations of $\mathcal{L}_{\text{match}}$ and $\mathcal{L}_{\text{reg}}$ are provided in the supplementary material.
\newcommand{\xmark}{\text{\texttimes}}
\newcommand{\cmark}{\textsf{\checkmark}}
\newcommand{\gap}{0.02\textwidth} 

\section{Experiments}
\label{sec:exp}

\subsection{Datasets}
\paragraph{KITTI dataset.}
KITTI~\cite{geiger2013vision} contains forward-facing ground-view images collected by a vehicle in Karlsruhe, Germany, covering urban, rural, and highway scenes. Following~\cite{shi2022beyond}, each ground image is paired with a 100 m$\times$100 m satellite patch at 0.20 m/pixel resolution, with the camera assumed to be located within the central 40 m$\times$40 m region. 
To simulate a realistic scenario without an orientation prior, we introduce a random orientation sampled uniformly from $[-\pi, \pi)$, following~\cite{lentsch2023slicematch,xia2023convolutional}.

\begin{table*}[t]  
    \caption{Position and orientation estimation errors on the VIGOR~\cite{zhu2021vigor} dataset. `Aligned' refers to the case where the ground image orientation is known. For unaligned images, the models estimate the full 360-degree. $^{\dagger}$ indicates the use of a two-step inference specific to the method in \cite{xia2025fg} to boost performance. Best in \textbf{bold}, second best \underline{underlined}.}
    \label{tab:vigor}
    \begin{centering}
    \begin{threeparttable}
    \resizebox{0.78\textwidth}{!}{
        \footnotesize{
        \begin{tabular}{lcccccccccc}
            \toprule
                &
                &
                & \multicolumn{4}{c}{Same-Area}
                & \multicolumn{4}{c}{Cross-Area} \\
            \cmidrule(lr){4-7} \cmidrule(lr){8-11}
                & &
                & \multicolumn{2}{c}{Position (m) ↓} 
                & \multicolumn{2}{c}{Orientation ($^{\circ}$) ↓}
                & \multicolumn{2}{c}{Position (m) ↓} 
                & \multicolumn{2}{c}{Orientation ($^{\circ}$) ↓} \\
                Method & Backbone & Aligned
                & Mean & Median 
                & Mean & Median 
                & Mean & Median 
                & Mean & Median \\
            \midrule
            MCC~\cite{xia2022visual} & VGG16 & \cmark
                & 6.94 & 3.64 
                & - & -
                & 9.05 & 5.14
                & - & - \\
            SliceMatch~\cite{lentsch2023slicematch} & VGG16 & \cmark
                & 5.18 & 2.58 
                & - & -
                & 5.53 & 2.55
                & - & - \\
            CCVPE~\cite{xia2023convolutional} & EfficientNet-B0 & \cmark
                & 3.60 & 1.36
                & - & -
                & 4.97 & 1.68
                & - & - \\
            DenseFlow~\cite{song2023learning} & ResNet18 & \cmark
                & 3.03 & \textbf{0.97}
                & - & -
                & 5.01 & 2.42
                & - & - \\
            HC-Net~\cite{wang2023fine} & EfficientNet-B0 & \cmark
                & 2.65 & 1.17
                & - & -
                & \underline{3.35} & 1.59
                & - & - \\
            FG2~\cite{xia2025fg} & DINOv2 & \cmark
                & \textbf{1.95} & \underline{1.08}
                & - & -
                & \textbf{2.41} & \textbf{1.37} 
                & - & - \\
            \rowcolor{gray!20} \textbf{VIRD (Ours)} & VGG16 & \cmark
                & 3.14 & 1.39 
                & - & -
                & 4.83 & 1.77
                & - & - \\
            \rowcolor{gray!20} \textbf{VIRD (Ours)} & EfficientNet-B0 & \cmark
                & \underline{2.57} & 1.14 
                & - & -
                & 3.85 & \underline{1.47}
                & - & - \\
            \midrule
            MCC~\cite{xia2022visual} & VGG16 & \xmark
                & 9.87 & 6.25 
                & 56.86 & 16.02
                & 12.66 & 9.55
                & 72.13 & 29.97 \\
            SliceMatch~\cite{lentsch2023slicematch} & VGG16 & \xmark
                & 8.41 & 5.07 
                & 28.43 & 5.15
                & 8.48 & 5.64
                & 26.20 & 5.18 \\
            CCVPE~\cite{xia2023convolutional} & EfficientNet-B0 & \xmark
                & 3.74 & \underline{1.42} 
                & 12.83 & 6.62
                & \underline{5.41} & \underline{1.89}
                & 27.78 & 13.58 \\
            DenseFlow~\cite{song2023learning} & ResNet18 & \xmark
                & 4.97 & 1.90
                & \underline{11.20} & 1.59
                & 7.67 & 3.67 
                & \underline{17.63} & 2.94 \\
            FG2~\cite{xia2025fg} & DINOv2 & \xmark
                & 8.95 & 7.32
                & 15.02 & 2.94
                & 10.02 & 8.14
                & 31.41 & 5.45 \\
            FG2$^{\dagger}$~\cite{xia2025fg} & DINOv2 & \xmark
                & 3.78 & 1.70
                & 12.63 & 1.44
                & 5.95 & 2.40
                & 28.41 & 2.20 \\
            \rowcolor{gray!20} \textbf{VIRD (Ours)} & VGG16 & \xmark
                & \underline{3.71} & 1.52
                & 12.45 & \underline{1.21}
                & 6.05 & 1.95
                & 24.55 & \underline{1.56} \\
            \rowcolor{gray!20} \textbf{VIRD (Ours)} & EfficientNet-B0 & \xmark
                & \textbf{3.00} & \textbf{1.20}
                & \textbf{9.52} & \textbf{0.96}
                & \textbf{4.61} & \textbf{1.55}
                & \textbf{16.50} & \textbf{1.17} \\
            \bottomrule
        \end{tabular}
        }}
    \end{threeparttable}
    \end{centering}
\end{table*}

\paragraph{VIGOR dataset.}
VIGOR~\cite{zhu2021vigor} consists of ground-view panoramas and corresponding satellite images from four U.S. cities. Each satellite image covers a 70 m$\times$70 m area. Ground panoramas are aligned so their vertical center corresponds to the north in satellite images. A satellite patch is labeled as positive if the ground camera lies within its central quarter region. Otherwise, it is semi-positive. We use only positive pairs, following~\cite{lentsch2023slicematch,xia2023convolutional}. The same-area and cross-area splits are adopted from~\cite{zhu2021vigor}, with corrected labels following~\cite{lentsch2023slicematch}. We evaluate the model under both known and unknown orientations, following~\cite{lentsch2023slicematch,xia2023convolutional}.

\subsection{Evaluation metrics}
We adopt the evaluation metrics from~\cite{lentsch2023slicematch}, reporting mean and median errors for position (in meters) and orientation (in degrees) across all datasets. 
For KITTI, we additionally compute recall following~\cite{lentsch2023slicematch, xia2023convolutional}, defined as the percentage of predictions within 1 m and 5 m in both lateral and longitudinal directions, and within $1^{\circ}$ and $5^{\circ}$ in orientation.

\subsection{Implementation details}
\label{sec:implementation}
Consistent with SliceMatch~\cite{lentsch2023slicematch} and CCVPE~\cite{xia2023convolutional}, we use a VGG16~\cite{simonyan2014very} and an EfficientNet-B0~\cite{tan2019efficientnet} backbone pretrained on ImageNet~\cite{deng2009imagenet}. 
Candidate poses are sampled on a uniform grid over the satellite image. For KITTI, we use a 5$\times$5 grid during training and 20$\times$20 at test time. For VIGOR, the grid is 7$\times$7 for training and 25$\times$25 for testing. The sampling stride determines the number of orientation candidates, with 16 candidates during training for both datasets, and 70 for KITTI and 80 for VIGOR in testing.
Training is performed for 10 epochs on an NVIDIA RTX A5000 GPU using the Adam optimizer~\cite{KingBa15}, with a learning rate of $10^{-4}$ and a batch size of 4. Further details are provided in the supplementary material.

\subsection{Comparison with state-of-the-art methods}

\paragraph{KITTI dataset.}
\cref{tab:kitti} compares the proposed method with the state-of-the-art methods on the KITTI dataset for VGG16 and EfficientNet-B0 backbones.
Our method consistently outperformed all baselines in both same-area and cross-area settings, demonstrating its effectiveness in omnidirectional cross-view pose estimation. With EfficientNet-B0, it reduced the median position and orientation errors by 50.7\% and 76.5\% in the cross-area setting.

These superior results stem from VIRD’s ability to explicitly bridge the viewpoint gap and integrate a robust coarse-to-fine pose estimation framework. Existing approaches suffer from inherent limitations in either representation or architecture. Geometry-based methods using projective transformations~\cite{shi2022beyond, song2023learning} and semantics-based methods~\cite{lentsch2023slicematch, xia2023convolutional} often leave spatial misalignment unresolved, thereby degrading pose estimation accuracy. HighlyAccurate~\cite{shi2022beyond}, an iterative optimization-based framework with a narrow search range, tends to converge to suboptimal solutions under unreliable orientation priors. 
Global descriptor-based methods such as SliceMatch~\cite{lentsch2023slicematch} remain limited by coarse matching accuracy.
Even FG2~\cite{xia2025fg}, which leverages height-aware 3D point selection to mitigate the viewpoint gap, degrades in orientation estimation due to its sparse point matching mechanism. 
In contrast, VIRD jointly resolves horizontal and vertical misalignment through dual-axis transformation and refines poses via a fine-grained regression after coarse global matching, achieving stable convergence and superior accuracy across metrics.

\begin{table}[t]
  \caption{Ablation study of dual-axis transformation on KITTI~\cite{geiger2013vision}.}
  \label{tab:pa_ablation}
  \centering
  \footnotesize
  \begin{threeparttable}
    \vspace{-2mm}
    \setlength{\tabcolsep}{4pt}
    \renewcommand{\arraystretch}{0.95}
    \begin{tabular}{lcccc}
      \toprule
      & \multicolumn{2}{c}{Pos. error (m) ↓} & \multicolumn{2}{c}{Orien. error ($^\circ$) ↓} \\
      & Mean & Med. & Mean & Med. \\
      \midrule
      Projective (S2G) & 14.37 & 10.59 & 37.92 & 3.84 \\
      Projective (G2S) & 17.10 & 15.20 & 47.62 & 5.44 \\
      Polar            & 15.26 & 11.75 & 39.92 & 4.00 \\
      \midrule
      Polar + PA (Ours)   & 13.92 &  9.76 & 35.17 & 3.44 \\
      Polar + CEPA (Ours) & \textbf{13.83} & \textbf{8.88} & \textbf{31.35} & \textbf{3.36} \\
      \bottomrule
    \end{tabular}
  \end{threeparttable}
\end{table}

\paragraph{VIGOR dataset.} 
\cref{tab:vigor} presents the evaluation results on the VIGOR dataset. With the EfficientNet-B0 backbone, our method reduced median position and orientation errors by 18.0\% and 46.8\% under the unaligned and cross-area settings, demonstrating strong generalization ability for pose estimation. Although FG2~\cite{xia2025fg} performs well in the aligned setting by leveraging 3D structural information via point matching, its accuracy drops sharply in the unaligned setting due to sparse point correspondences. This limitation reduces scalability in real-world environments where reliable orientation priors are often unavailable. To alleviate this issue, FG2 introduces a two-step variant (denoted as $\text{FG2}^{\dagger}$ in \cref{tab:vigor}) that sequentially estimates orientation and refines pose, yet it still underperforms compared with VIRD.

\begin{figure*}[ht]
    \begin{subfigure}{\textwidth}
        \centering
        \includegraphics[width=\textwidth]{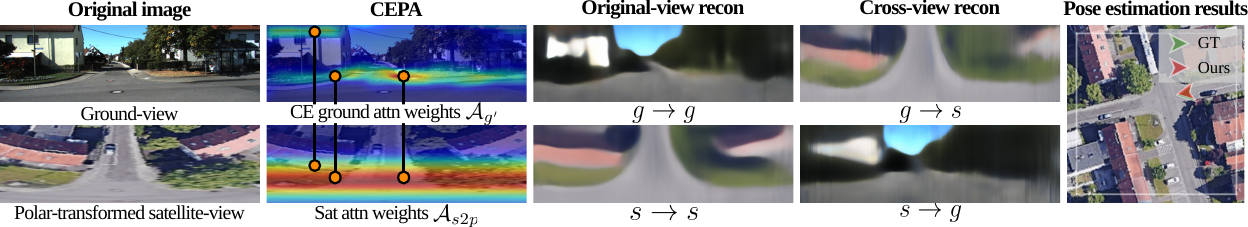}
        \caption{Example 1}
        \label{fig:figure5_a}
    \end{subfigure}
    \begin{subfigure}{\textwidth}
        \centering
        \includegraphics[width=\textwidth]{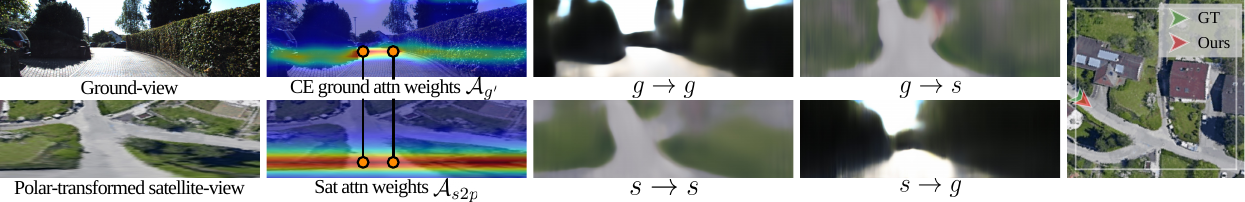}
        \caption{Example 2}
        \label{fig:figure5_b}
    \end{subfigure}
    \caption{Visualization of CEPA's attention weights, view reconstruction, and pose estimation results from VIRD.}
    \label{fig:figure5}
\end{figure*} 

\subsection{Ablation study}
All ablations were conducted on the KITTI dataset using a VGG16 backbone in the cross-area evaluation setting. 

\begin{table}[t]
  \caption{Ablation study of model components on KITTI~\cite{geiger2013vision}.}
  \label{tab:module_ablation}
  \centering
  \footnotesize
  \begin{threeparttable}
    \vspace{-2mm}
    \setlength{\tabcolsep}{4pt}
    \renewcommand{\arraystretch}{0.95}
    \begin{tabular}{lcccc}
      \toprule
      & \multicolumn{2}{c}{Pos. error (m) ↓} & \multicolumn{2}{c}{Orien. error ($^\circ$) ↓} \\
      & Mean & Med. & Mean & Med. \\
      \midrule
      Polar + CEPA          & 13.83 & 8.88 & 31.35 & 3.36 \\
      \midrule
      + $\mathcal{L}_{\text{origin}}$     & 13.34 & 8.29 & 28.26 & 3.31 \\
      + $\mathcal{L}_{\text{cross}}$     & 12.84 & 8.10 & 26.30 & 3.21 \\
      + $\mathcal{L}_{\text{origin}}$ + $\mathcal{L}_{\text{cross}}$     & 12.72 & 7.90 & \textbf{21.91} & 3.05 \\
      + $\mathcal{L}_{\text{origin}}$ + $\mathcal{L}_{\text{cross}}$+ $\mathcal{L}_{\text{reg}}$         & \textbf{12.30} & \textbf{7.05} & 25.10 & \textbf{2.22} \\
      \bottomrule
    \end{tabular}
  \end{threeparttable}
\end{table}

\begin{figure}[ht]
    \centering
    \includegraphics[width=0.58\columnwidth]{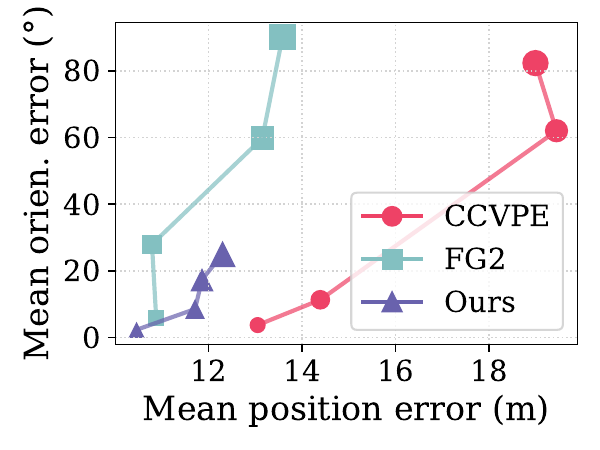}
    \caption{
    Ablation study of robustness to orientation noise on KITTI~\cite{geiger2013vision}. Larger markers denote higher noise levels ($\pm10^\circ$, $\pm60^\circ$, $\pm120^\circ$, and $\pm180^\circ$). 
    }
    \label{fig:figure6}
\end{figure}

\paragraph{Dual-axis transformation.}
As shown in \cref{tab:pa_ablation}, we evaluated the effectiveness of dual-axis transformation. 
The baselines include satellite-to-ground projective transformation~\cite{shi2022beyond} (S2G), ground-to-satellite projective transformation~\cite{shi2023boosting} (G2S), and polar transformation without attention (Polar).
Incorporating positional attention (Polar + PA) outperformed all baselines, achieving median errors of 9.76 m and 3.44$^\circ$ for position and orientation, respectively. 
Adding contextual information (Polar + CEPA) further reduced the median position error to 8.88 m. 
These results highlight that the dual-axis transformation is a powerful alternative to traditional geometry-based approaches. 
\cref{fig:figure5} (second column) visualizes CEPA's attention weights, demonstrating its ability to capture vertical correspondences between structures in the ground view (\eg, roofs, roads) and the corresponding vertical positions in the satellite view. 

\paragraph{View-reconstruction loss.}
\cref{tab:module_ablation} evaluates the impact of the view-reconstruction loss. Incorporating both original and cross-view reconstruction consistently improves performance compared with CEPA without reconstruction. The most notable improvement is a 39.1\% reduction in mean orientation error when applying both reconstruction losses. This result highlights the importance of capturing vertical structures, particularly in ambiguous scenarios where similar road shapes can lead to reversed orientation predictions. 
 As illustrated in the third and fourth columns of \cref{fig:figure5}, the trained descriptors successfully reconstruct road layouts and approximate building shapes while filtering out objects not shared across views such as cars and side roads. These findings demonstrate that the view-reconstruction loss promotes view-invariant representations by encouraging the model to focus on cross-view consistent structures.

\paragraph{Regression module.}

The regression module refines the predicted coarse pose to improve pose estimation precision. As shown in \cref{tab:module_ablation}, it reduces the median position and orientation errors by 0.85 m and 0.83$^{\circ}$, respectively, compared with the model without regression. Although the regression module slightly increases the mean orientation error, this effect arises from higher directional precision: when coarse matching predicts the opposite direction, the refined orientation becomes farther from the ground truth, while predictions close to the correct direction become more precise.

\paragraph{Robustness to orientation noise.}
\cref{fig:figure6} shows the mean position and orientation errors of CCVPE~\cite{xia2023convolutional}, FG2~\cite{xia2025fg}, and our VIRD under different levels of orientation noise ($\pm10^\circ$, $\pm60^\circ$, $\pm120^\circ$, and $\pm180^\circ$). Because orientation noise simultaneously affects both position and orientation estimates, we jointly visualize them to capture their coupled behavior.
As orientation noise increases, the errors become larger, and greater distances between markers indicate more severe performance degradation. VIRD consistently maintains low errors across all noise levels, demonstrating strong robustness to orientation noise compared with CCVPE and FG2. 

\section{Conclusion}
This paper presented VIRD, a novel cross-view pose estimation method that bridges the ground–satellite viewpoint gap by constructing view-invariant representations. 
To ensure view invariance, VIRD introduces a dual-axis transformation and a view-reconstruction loss. 
Experiments on the KITTI and VIGOR datasets demonstrate the superiority of the proposed method. 
Future work will extend this approach to more challenging 6-DoF pose estimation tasks for both on the ground and airborne.

{
    \small
    \bibliographystyle{ieeenat_fullname}
    \bibliography{main}
}

\clearpage
\setcounter{page}{1}
\maketitlesupplementary

\makeatletter
\renewcommand{\theHsection}{supp.\thesection}
\makeatother

\setcounter{section}{0} 
\setcounter{table}{0} 
\setcounter{figure}{0} 

\renewcommand{\thesection}{\Alph{section}}
\renewcommand{\thetable}{\Alph{table}}
\renewcommand{\thefigure}{\Alph{figure}}

\section*{Table of Contents}
\hypersetup{linkcolor=black}
\setlength{\cftbeforesecskip}{0.6pt}

\addtocontents{toc}{\protect\setcounter{tocdepth}{1}}

\makeatletter
\@starttoc{toc}
\makeatother

\hypersetup{linkcolor=blue}
\vspace{1em}

Unless otherwise specified, all experiments are conducted on the KITTI~\cite{geiger2013vision} dataset under the cross-area setting.

\section{Context-Enhanced Positional Attention}
\label{sec:suppl_cepa}
\subsection{Pseudocode}
\begin{algorithm}[H]
\footnotesize
\caption{Pseudocode for CEPA}
\begin{algorithmic}[1]

\State \textbf{Input:}
\State \quad Query positional encoding $P_a$ \hfill \# $(H_Q, d_p)$
\State \quad Key positional encoding $P_v$ \hfill \# $(H_K, d_p)$
\State \quad Ground attention weights $A_g$ \hfill \# $(H_Q, H_K)$
\State \quad Ground feature $F_g$ \hfill \# $(B, C, H_K, W_g)$

\State \textbf{Output:}
\State \quad Positional attention weights $A_v$ \hfill \# $(H_Q, H_K)$
\State \quad Context-enhanced attention weights $A_{g'}$ \hfill \# $(B, H_Q, H_K, W_g)$

\vspace{4pt}
\Function{cal\_positional\_attn\_weights}{$P_a, P_v$}
    \State $Q \leftarrow \text{linear\_projection}(P_a)$
    \State $K \leftarrow \text{linear\_projection}(P_v)$
    \State $A_v \leftarrow (QK^\top)/\sqrt{d_k}$
    \State $A_v \leftarrow \text{softmax}(A_v,\; \text{dim}=-1)$
    \State \Return $A_v$
\EndFunction

\vspace{4pt}
\Function{context\_enhancement}{$A_g, F_g$}

    \State $A_g' \leftarrow \text{reshape}(A_g,\; 1, H_Q, 1, H_K, 1)$
    \State $A_g' \leftarrow \text{repeat}(A_g',\; B, 1, 1, 1, W_g)$ \hfill \# $(B, H_Q, 1, H_K, W_g)$\\

    \State $F_g' \leftarrow \text{reshape}(F_g,\; B, 1, C, H_K, W_g)$
    \State $F_g' \leftarrow \text{repeat}(F_g',\; 1, H_Q, 1, 1, 1)$ \hfill \# $(B, H_Q, C, H_K, W_g)$\\

    \State $\text{feat} \leftarrow \text{concat}(A_g', F_g',\; \text{dim}=2)$ \hfill \# $(B, H_Q, C{+}1, H_K, W_g)$

    \State $\text{feat} \leftarrow \text{reshape}(\text{feat},\; B\!\cdot\!H_Q,\; C{+}1,\; H_K,\; W_g)$
    \State $\text{feat} \leftarrow \text{conv2d}(\text{feat})$ \hfill \# $(B\!\cdot\!H_Q, 1, H_K, W_g)$
    \State $\text{feat} \leftarrow \text{reshape}(\text{feat},\; B, H_Q, H_K, W_g)$

    \State $\hat{A} \leftarrow \text{softmax}(\text{feat},\; \text{dim}=-2)$
    \State $A_{g'} \leftarrow \hat{A} + A_g'$ \hfill \# after broadcasting
    \State \Return $A_{g'}$

\EndFunction

\end{algorithmic}
\end{algorithm}

\begin{algorithm}[H]
\footnotesize
\caption{Pseudocode for vertical feature transformation}
\begin{algorithmic}[1]

\State \textbf{Input:}
\State \quad Shared virtual positional encoding $P_a$ \hfill \# $(H_Q, d_p)$
\State \quad Ground positional encoding $P_g$ \hfill \# $(H_K, d_p)$
\State \quad Satellite positional encoding $P_{s2p}$ \hfill \# $(H_K, d_p)$
\State \quad Ground feature $F_g$ \hfill \# $(B, C, H_K, W_g)$
\State \quad Polar-transformed satellite feature $F_{s2p}$ \hfill \# $(B, C, H_K, W_s)$

\State \textbf{Output:}
\State \quad Vertically-transformed ground feature $F_{g'}$ \hfill \# $(B, C, H_Q, W_g)$
\State \quad Vertically-transformed satellite feature $F_{s2p'}$ \hfill \# $(B, C, H_Q, W_s)$

\vspace{4pt}
\Function{vertical\_transform}{$P_a, P_g, P_{s2p}, F_g, F_{s2p}$}

    \State $A_g \leftarrow \Call{cal\_positional\_attn\_weights}{P_a, P_g}$
    \State $A_{s2p} \leftarrow \Call{cal\_positional\_attn\_weights}{P_a, P_{s2p}}$
    \State $A_{g'} \leftarrow \Call{context\_enhancement}{A_g, F_g}$

    \State $F_{g'}[b,c,h_q,w] \leftarrow \sum_{k} A_{g'}[b,h_q,h_k,w]\;F_g[b,c,h_k,w]$
    \State $F_{s2p'}[b,c,h_q,w] \leftarrow \sum_{k} A_{s2p}[h_q,h_k]\;F_{s2p}[b,c,h_k,w]$

    \State \Return $F_{g'}, F_{s2p'}$

\EndFunction

\end{algorithmic}
\end{algorithm}

\subsection{CEPA vs. content-based attention}
\begin{figure}[t]
    \centering
    \includegraphics[width=\columnwidth]{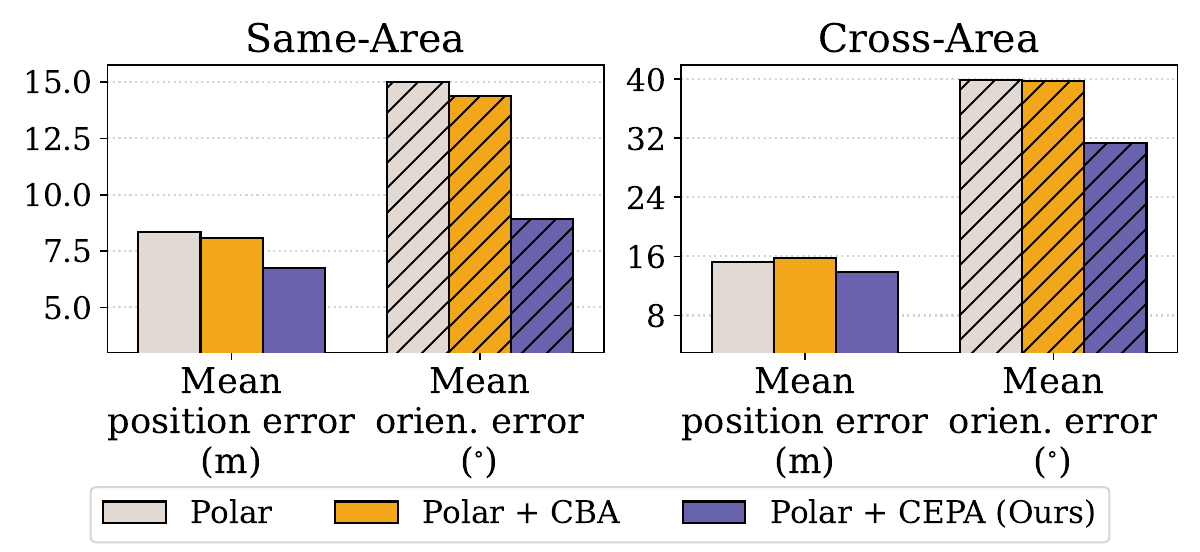}
    \caption{
    Comparison of content-based attention (CBA) and context-enhanced positional attention (CEPA) on KITTI~\cite{geiger2013vision}.}
    \label{fig:figureA}
\end{figure}

\cref{fig:figureA} compares the proposed context-enhanced positional attention (CEPA) with a content-based attention (CBA) baseline. CBA follows the conventional formulation in which attention weights are computed directly from input content, such as image features~\cite{schmidt2023bridging}. We implement CBA using the self- and cross-attention layers from SliceMatch~\cite{lentsch2023slicematch} and apply them with a polar transformation (Polar + CBA). CBA yields a moderate gain in the same-area setting, where training and test scenes share similar visual patterns. However, its performance degrades in the cross-area setting. Because CBA derives attention weights from image content, it overfits to scene-specific semantics and fails to capture the spatial correspondence across unseen environments.
In contrast, CEPA consistently achieves superior performance in both same- and cross-area evaluations. It first derives attention weights from positional cues and then refines them using ground context, allowing the model to more effectively focus on spatial correspondence rather than appearance. These results demonstrate that explicitly modeling positional correspondence is crucial for robust cross-view pose estimation.

\subsection{Positional encoding types}
\cref{tab:pa_positional} compares different types of positional encodings, including sinusoidal, sinusoidal-learnable, learnable, and 2D sinusoidal variants. The sinusoidal encoding is fixed and follows the formulation of~\cite{vaswani2017attention}. The sinusoidal-learnable variant applies a nonlinear projection to the fixed sinusoidal encoding. In contrast, the learnable encoding is initialized as a fully trainable tensor without relying on any predefined positional pattern. The 2D sinusoidal variant extends positional information to both horizontal and vertical directions in the ground view. 

Among these variants, the fixed 1D sinusoidal encoding (Polar + PA in the main paper’s \cref{tab:pa_ablation}) achieves the best overall performance, showing the lowest position error and competitive orientation accuracy. These results indicate that vertical transformations across views can be learned more reliably when positional encodings are fixed and 1D. Notably, the comparison with the 2D sinusoidal variant suggests that encoding both horizontal and vertical positional information may introduce ambiguity in establishing spatial correspondence. This may occur because the standard positional attention mechanism relies solely on positional information, which might not adequately account for the complexity of vertical structures in ground-view images. In contrast, assuming a consistent vertical transformation along the horizontal direction provides a more stable inductive bias that facilitates reliable spatial correspondence. This finding further highlights the strength of the proposed CEPA, which uses fixed 1D positional encoding to model shared vertical transformation while adaptively capturing horizontal variations in the ground view through context enhancement.

\begin{table}[t]
    \caption{Positional attention performance with various positional encoding variants on KITTI~\cite{geiger2013vision}.}
    \label{tab:pa_positional}
  \centering
  \footnotesize
  \begin{threeparttable}
    \vspace{-2mm}
    \setlength{\tabcolsep}{4pt}
    \renewcommand{\arraystretch}{0.95}
        \begin{tabular}{lcccc}
            \toprule
            & \multicolumn{2}{c}{Pos. error (m) ↓} 
            & \multicolumn{2}{c}{Orien. error ($^{\circ}$) ↓} \\
            &  Mean & Med. & Mean & Med. \\
            \midrule
            Sinusoidal pe
            & \textbf{13.92} & \textbf{9.76} 
            & 35.17 & 3.44 \\
            Sinusoidal learnable pe
            & 14.28 & 10.63 
            & 33.82 & \textbf{3.41} \\
            Learnable pe
            & 15.23 & 12.06 
            & 36.23 & 4.04 \\
            2D sinusoidal pe
            & 14.68 & 10.31 
            & \textbf{33.57} & 3.57 \\
            \bottomrule
        \end{tabular}
  \end{threeparttable}
\end{table}

\section{View-Reconstruction Loss}
\label{sec:suppl_view_recon}
\subsection{Effect of view-reconstruction on CEPA}
\begin{figure*}[t]
    \centering
    \includegraphics[width=0.9\textwidth]{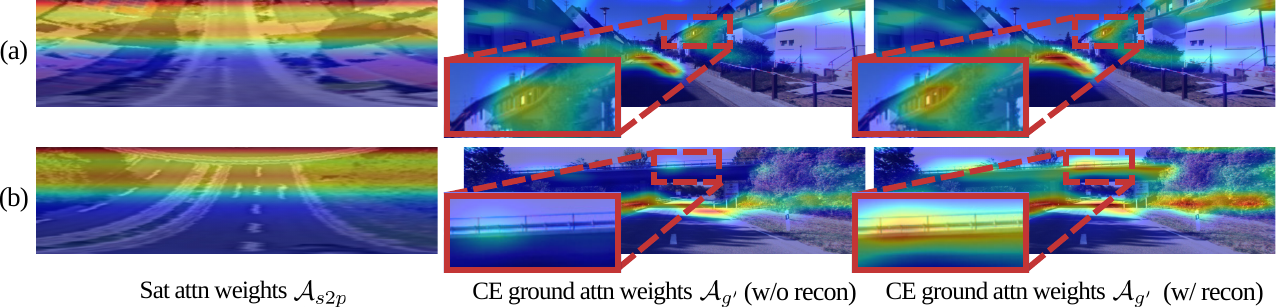}
    \caption{
    Effect of the view-reconstruction loss on CEPA. The first row presents example (a), and the second row presents example (b). For each example, the satellite attention $\mathcal{A}_{s2p}$ and the context-enhanced ground attention $\mathcal{A}_{g'}$ are visualized at the same vertical position in the shared virtual positional encoding space. The view-reconstruction loss stabilizes the learning of vertical transformations in CEPA, yielding stronger roof activations in example (a) and successful detection of overpass structures in example (b). Regions with stronger activations are highlighted in red.}
    \label{fig:figureB}
\end{figure*}

This section analyzes how the view-reconstruction loss influences the learning of CEPA.
During training, the view-reconstruction loss encourages the ground and satellite descriptors to reconstruct both their original and cross views. 
This supervision enables the descriptors to embed structural cues more effectively, not only from regions with small viewpoint gaps, such as road surfaces, but also from structures with large viewpoint gaps, including tall buildings and other vertically dominant elements. By promoting the preservation of meaningful information even in these challenging regions, the view-reconstruction loss allows CEPA to attend more effectively to vertical structures and ultimately learn more stable vertical transformations between the cross views.
As shown in \cref{fig:figureB}, the visualization of the context-enhanced ground attention weights highlights this effect.
In example (a), CEPA with the view-reconstruction loss exhibits stronger activations around roof regions, indicating improved vertical correspondence.
In example (b), the model successfully captures the overpass structure that is missed when the view-reconstruction loss is removed.

\begin{table}[t]
    \caption{Comparison of different reconstruction loss types on KITTI~\cite{geiger2013vision}.}
    \label{tab:view_recon_ablation}
  \centering
  \footnotesize
  \begin{threeparttable}
    \vspace{-2mm}
    \setlength{\tabcolsep}{4pt}
    \renewcommand{\arraystretch}{0.95}
        \begin{tabular}{lcccc}
            \toprule
            & \multicolumn{2}{c}{Pos. error (m) ↓} 
            & \multicolumn{2}{c}{Orien. error ($^{\circ}$) ↓} \\
            &  Mean & Med. & Mean & Med. \\
            \midrule
            $\ell_1$
            & \textbf{12.72 }& \textbf{7.90 }
            & \textbf{21.91} & \textbf{3.05} \\
            $\ell_2$
            & 13.14 & 8.30 
            & 25.74 & 3.31 \\
            SSIM
            & 12.92 & 8.00
            & 25.23 & 3.34 \\
            Percep
            & 14.00 & 9.78 
            & 29.55 & 3.31 \\
            \bottomrule
        \end{tabular}
  \end{threeparttable}
\end{table}

\subsection{Reconstruction loss types}
Several loss functions for view reconstruction are evaluated, including $\ell_1$, $\ell_2$, SSIM, and perceptual loss, as summarized in \cref{tab:view_recon_ablation}. Among them, the $\ell_1$ loss achieves the best performance, yielding a mean position error of 12.72 m and a mean orientation error of 21.91$^\circ$.

The goal of view reconstruction in our framework is not to generate perceptually realistic images, but to provide stable structural supervision that promotes view-invariant descriptors. $\ell_2$ loss is less suitable because it strongly penalizes large pixel differences that naturally arise between ground and satellite views, leading to unstable gradients. SSIM focuses on luminance and contrast consistency, which are not reliably shared across cross views, thereby weakening the geometric cues needed for spatial correspondence. Perceptual loss is also ineffective: it compares high-level semantic features extracted by networks trained on perspective natural images, but such semantics differ substantially between ground and satellite views, producing inconsistent gradients that degrade geometric structure. In contrast, the $\ell_1$ loss provides uniformly weighted, geometry-preserving supervision that avoids overfitting to appearance or semantics. This property makes the $\ell_1$ loss the most stable and effective choice for enforcing view invariance in cross-view matching.

\section{Matching and Regression Modules}
\label{sec:suppl_matching_regression}

\subsection{Shifting and cropping of satellite descriptor}
\label{sec:shift_and_crop}
Given a candidate pose $\Pose{c} = (x_c, y_c, \theta_c)$, a satellite descriptor $D_{s2p}^{(x_c, y_c)} \in \mathbb{R}^{K_s}$ is constructed via dual-axis transformation and vertical directional encoding. This descriptor is initially aligned such that its central vertical line corresponds to the east, i.e., $\theta = 0$. To align it with the orientation $\theta_c$, the descriptor is cyclically shifted along the horizontal axis by $\frac{\theta_c}{2\pi} \cdot W_s$, following prior work~\cite{xia2023convolutional}. After alignment, the descriptor is center-cropped to match the size of the ground descriptor $D_g \in \mathbb{R}^{K_g}$, yielding the orientation-aligned satellite descriptor $D_{s2p}^{\Pose{c}} \in \mathbb{R}^{K_g}$. Similarly, the polar-transformed satellite image $I_{s2p}^{\Pose{c}}$, which is used in the view-reconstruction loss, is shifted and cropped in the same manner to ensure spatial alignment with $I_g$.

\subsection{Regression network architecture}
The regression module first reshapes the 1D descriptor difference into a 2D representation with channel and width dimensions, then passes it through a series of 1D convolutional layers to obtain a high-dimensional embedding. The resulting feature is flattened and concatenated with the coarse pose. MLPs are applied to predict the pose residual $\Delta \Pose{} = (\Delta x, \Delta y, \Delta \theta)$ relative to the ground truth. The predicted output is scaled to operate within a predefined search range.

\subsection{Training details}
\paragraph{Matching loss.}
To train discriminative descriptors, we adopt the InfoNCE loss~\cite{oord2018representation}, encouraging high similarity at the ground-truth pose $\Pose{}^*$ and penalizing $n$-th non-matching pose $\Pose{n}$, following~\cite{lentsch2023slicematch, xia2023convolutional}:
\begin{equation}
\mathcal{L}_{\text{match}} = -\log \left( \frac{\exp(S^{\Pose{}^*}/\tau)}{\sum_{n=1}^N \exp(S^{\Pose{n}}/\tau) + \exp(S^{\Pose{}^*}/\tau)} \right),
\end{equation}
where $S^{\Pose{}^*}$ and $S^{\Pose{n}}$ are similarity scores at the ground-truth pose and the $n$-th non-matching pose among $N$ samples, respectively. $\tau$ is a temperature scaling parameter.

\paragraph{Regression loss.}
The regression module is trained using the $\ell_2\text{-loss}$ computed over pose residuals:
\begin{equation}
    \mathcal{L}_{\text{reg}} = \beta ( \|\Delta x - \Delta x^*\|_2 + \|\Delta y - \Delta y^*\|_2 + \|\Delta \theta - \Delta \theta^*\|_2 ), 
\label{eq:regression_loss}
\end{equation}
where $(\Delta x^*, \Delta y^*, \Delta \theta^*)$ denote the ground-truth pose residuals, and $\beta$ is a weighting coefficient.

\paragraph{Regression training.} 
To train the regression module, $N_r$ coarse poses are randomly sampled around the ground-truth pose within predefined spatial and angular ranges, where reliable refinement is feasible. For each sampled pose $\Pose{r}$, the satellite descriptor $D_{s2p}^{\Pose{r}}$ is computed using the same shift-and-crop mechanism described in \cref{sec:shift_and_crop}. The regression loss for each sampled pose is computed as described in \cref{eq:regression_loss}, and the final loss is obtained by averaging over all $N_r$ samples.

\section{Additional Implementation Details}
\label{sec:suppl_implementation}
We provide additional implementation details in \cref{sec:implementation}.
The spatial size of $I_g$ is 256$\times$1024 for KITTI and 320$\times$640 for VIGOR, and $I_s$ is 512$\times$512 for both datasets. 
The temperature parameter $\tau$ is set to 0.05. 
The loss coefficients $\alpha_1$, $\alpha_2$, and $\beta$ are set to 1, 10, and 5, respectively. The polar transform is configured with $r_{\min}$ and $r_{\max}$ set to 6 m and 40 m for KITTI~\cite{geiger2013vision}, and to 0 m and 30 m for VIGOR~\cite{zhu2021vigor}, respectively. For the CEPA module, the shared virtual axis height $H_Q$ is set equal to the ground feature height $H$, which is 16 and 20 for KITTI and VIGOR, respectively. To improve computational efficiency, the number of channels is reduced by a factor of four after feature extraction with the VGG16 backbone. The feature extractor weights are shared when using the VGG16 backbone; in all other settings, weights are not shared. Zero-padding is used as the default setting. For panorama images, circular padding is applied along the horizontal axis. 
For regression, the search range is limited to $\pm$4 m, $\pm$4 m, and $\pm$3.6$^{\circ}$ in $x$, $y$, and $\theta$ dimensions, respectively. 

\section{Applicability to Pinhole Camera Models}
\label{sec:suppl_pinhole}
Following prior work~\cite{xia2023convolutional}, we assume a cylindrical projection for all cameras. This assumption is not strictly valid for pinhole cameras, meaning that the polar-transformed satellite image does not perfectly correspond to the horizontal axis of a pinhole ground image. Despite this mismatch, the polar transformation remains effective for reducing the viewpoint gap. As shown in \cref{tab:pinhole}, models trained with raw pinhole images and those trained with a cylindrical projection applied to the same images yield nearly identical performance. This result indicates that the projection mismatch has negligible influence on accuracy, demonstrating that VIRD can be applied to pinhole cameras without difficulty.

\begin{table}[t]
    \caption{Impact of cylindrical projection on pinhole camera on KITTI~\cite{geiger2013vision}.}
    \label{tab:pinhole}
  \centering
  \footnotesize
  \begin{threeparttable}
    \vspace{-2mm}
    \setlength{\tabcolsep}{4pt}
    \renewcommand{\arraystretch}{0.95}
        \begin{tabular}{lcccc}
            \toprule
            Cylindrical
            & \multicolumn{2}{c}{Pos. error (m) ↓} 
            & \multicolumn{2}{c}{Orien. error ($^{\circ}$) ↓} \\
            projection
            &  Mean & Med. & Mean & Med. \\
            \midrule
            \xmark
            & 15.26 & 11.75 
            & 39.92 & \textbf{4.00} \\
            \cmark
            & \textbf{15.10} & \textbf{11.62} 
            & \textbf{38.97} & 4.13 \\
            \bottomrule
        \end{tabular}
  \end{threeparttable}
\end{table}

\section{Additional Ablation Studies}
\label{sec:suppl_additional_abl}

\subsection{Hyperparameters}
\begin{table}[t]
    \caption{Ablation study of the temperature parameter on KITTI~\cite{geiger2013vision}.}
    \label{tab:temp_kitti_ablation}
  \centering
  \footnotesize
  \begin{threeparttable}
    \vspace{-2mm}
    \setlength{\tabcolsep}{4pt}
    \renewcommand{\arraystretch}{0.95}
        \begin{tabular}{ccccc}
            \toprule
            & \multicolumn{2}{c}{Pos. error (m) ↓} 
            & \multicolumn{2}{c}{Orien. error ($^{\circ}$) ↓} \\
            $\tau$ 
            &  Mean & Med. & Mean & Med. \\
            \midrule
            0.1
            & 16.09 & 13.21 
            & \textbf{38.54} & 4.34 \\
            0.05
            & \textbf{15.26} & \textbf{11.75} 
            & 39.92 & \textbf{4.00} \\
            0.01
            & 17.18 & 14.93 
            & 53.59 & 6.17 \\
            \bottomrule
        \end{tabular}
  \end{threeparttable}
\end{table}

\paragraph{Temperature}
The temperature parameter $\tau$ is evaluated using a baseline model that includes only the polar transformation, excluding the proposed attention and reconstruction modules. \cref{tab:temp_kitti_ablation} shows that the best performance is achieved at 0.05.

\paragraph{Loss coefficients}
\cref{tab:coe_kitti_ablation} presents the results of ablation experiments on the loss coefficients. The best performance is achieved when the loss coefficients for original-view reconstruction ($\alpha_1$), cross-view reconstruction ($\alpha_2$), and regression ($\beta$) are set to 1, 10, and 5, respectively. The matching loss coefficient is set to its default value of 1, and thus no ablation was performed for this parameter.

\paragraph{Regression search range}
\cref{tab:search_range_kitti_ablation} reports the effect of varying the search range for the regression module. The model achieves high average recall accuracy when the search range is limited to $\pm4$ m in translation and $\pm3.6^\circ$ in orientation. This setting performs comparably to the tighter configuration of $\pm2$ m and $\pm1.8^\circ$, while significantly outperforming broader ranges such as $\pm6$ m. These results demonstrate that the optimal search range for the regression module lies within a moderate range, where it can more precisely focus on local feature differences and more effectively correct residual pose errors.

\begin{table}[t]
    \caption{Ablation study of the loss coefficients on KITTI~\cite{geiger2013vision}.}
    \label{tab:coe_kitti_ablation}
  \centering
  \footnotesize
  \begin{threeparttable}
    \vspace{-2mm}
    \setlength{\tabcolsep}{4pt}
    \renewcommand{\arraystretch}{0.95}
        \begin{tabular}{ccccccc}
            \toprule
            & &
            & \multicolumn{2}{c}{Pos. error (m) ↓} 
            & \multicolumn{2}{c}{Orien. error ($^{\circ}$) ↓} \\
            $\alpha_1$ & $\alpha_2$ & $\beta$ 
            &  Mean & Med. & Mean & Med. \\
            \midrule
            0.1 & 0 & 0
            & 13.43 & 8.54
            & 30.08 & \textbf{3.30} \\
            1 & 0 & 0
            & \textbf{13.34} & \textbf{8.29} 
            & 28.26 & 3.31 \\
            5 & 0 & 0
            & 13.90 & 9.04 
            & \textbf{27.01} & 3.34 \\
            \midrule
            1 & 5 & 0
            & \textbf{12.61} & \textbf{7.64} 
            & 25.72 & 3.16 \\
            1 & 10 & 0
            & 12.72 & 7.90 
            & \textbf{21.91} & \textbf{3.05} \\
            1 & 20 & 0
            & 13.25 & 8.39 
            & 24.13 & 3.54 \\
            \midrule
            1 & 10 & 1
            & 12.45 & 7.48 
            & \textbf{23.59} & 2.47 \\
            1 & 10 & 5
            & \textbf{12.30} & \textbf{7.05} 
            & 25.10 & \textbf{2.22} \\
            \bottomrule
        \end{tabular}
  \end{threeparttable}
\end{table}
\begin{table}[t]
    \caption{Ablation study of the regression search range on KITTI~\cite{geiger2013vision}.}
    \label{tab:search_range_kitti_ablation}
  \centering
  \footnotesize
  \begin{threeparttable}
    \vspace{-2mm}
    \setlength{\tabcolsep}{4pt}
    \renewcommand{\arraystretch}{0.95}
        \begin{tabular}{ccccc}
            \toprule
            & 
            & \multicolumn{1}{c}{Lat. (\%) ↑} 
            & \multicolumn{1}{c}{Lon. (\%) ↑} 
            & \multicolumn{1}{c}{Orien. (\%) ↑} \\
            Range & Area
            &  R@1m & R@1m & R@1$^{\circ}$ \\
            \midrule
            $\pm 2$ m, $\pm 2$ m, $\pm 1.8^{\circ}$ & Same
            & 79.30 & 29.21 & 44.16 \\
            $\pm 4$ m, $\pm 4$ m, $\pm 3.6^{\circ}$ & Same
            & \textbf{79.46} & \textbf{31.65} & 49.32 \\
            $\pm 6$ m, $\pm 6$ m, $\pm 4.8^{\circ}$ & Same
            & 74.77 & 28.49 & \textbf{51.63} \\
            \midrule
            $\pm 2$ m, $\pm 2$ m, $\pm 1.8^{\circ}$ & Cross
            & \textbf{44.37} & \textbf{12.98} & 26.81 \\
            $\pm 4$ m, $\pm 4$ m, $\pm 3.6^{\circ}$ & Cross
            & 43.61 & 12.88 & \textbf{27.65} \\
            $\pm 6$ m, $\pm 6$ m, $\pm 4.8^{\circ}$ & Cross
            & 40.20 & 12.11 & 27.06 \\
            \bottomrule
        \end{tabular}
  \end{threeparttable}
\end{table}

\begin{figure}[t!]
    \centering
    \includegraphics[width=0.9\columnwidth]{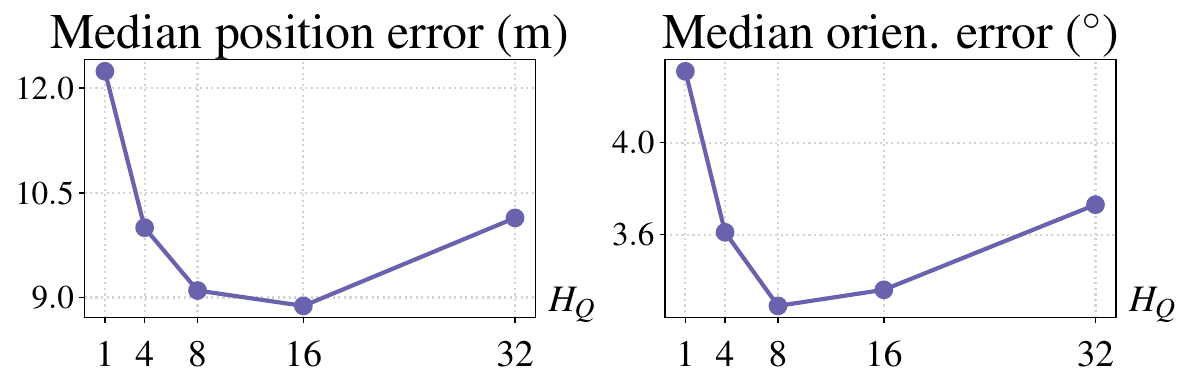}
    \caption{
    Ablation study of $H_Q$ on KITTI~\cite{geiger2013vision}.
    }
    \label{fig:figureC}
\end{figure}

\subsection{Impact of shared virtual axis height}
As shown in \cref{fig:figureC}, we conduct an ablation study on the height of the shared virtual axis $H_Q$, which defines the resolution of the shared coordinate system in the CEPA module. Performance peaks around $H_Q = 16$, demonstrating that this resolution is sufficient to capture vertical correspondences between cross views. Notably, performance drops markedly at $H_Q = 1$, where vertical information is simply collapsed into a single dimension without establishing vertical correspondence. This result supports the necessity of the shared virtual axis for resolving vertical misalignment. On the other hand, increasing $H_Q$ to 32 leads to degraded performance, suggesting that excessively high resolution may introduce redundancy that hinders stable learning.

\begin{figure}[t]
    \centering
    \includegraphics[width=\columnwidth]{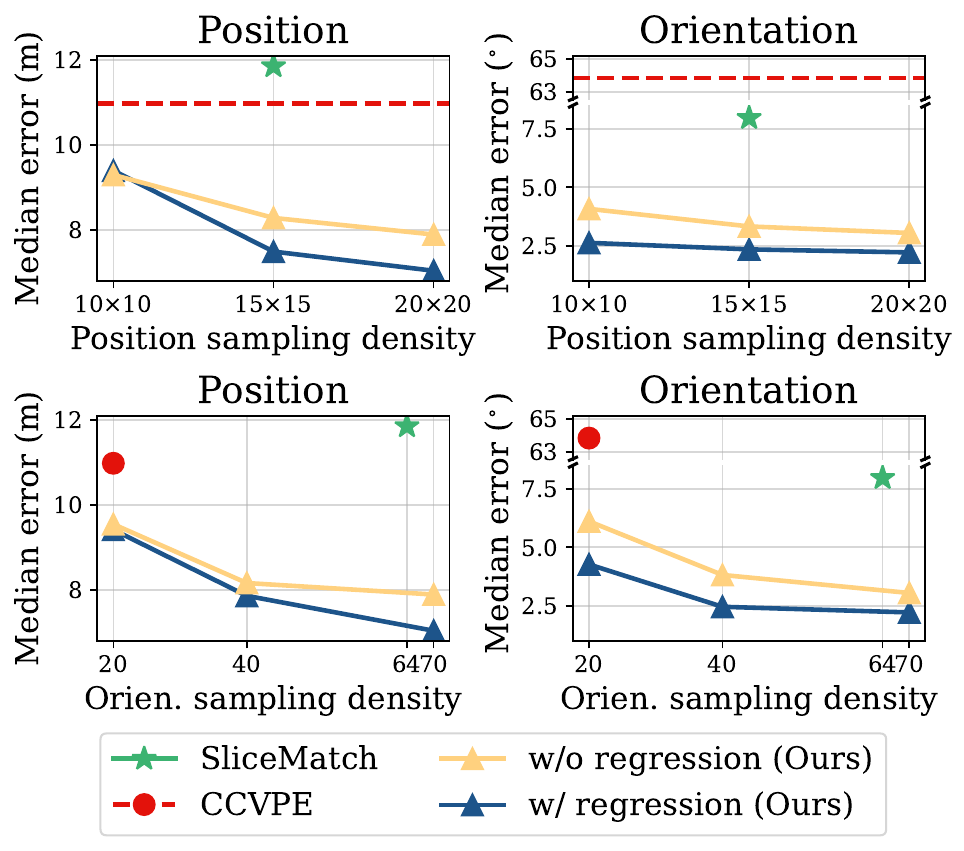}
    \caption{Ablation study of candidate pose sampling density, compared with SliceMatch~\cite{lentsch2023slicematch} and CCVPE~\cite{xia2023convolutional}. For the proposed method, the number of orientation candidates is fixed to 70 during position sampling, and the position grid size is fixed to 20$\times$20 during orientation sampling. SliceMatch performs matching only on 15$\times$15$\times$64 candidate poses. For CCVPE, position prediction is performed at the original resolution, while orientation prediction first matches 20 candidates and then performs regression. Because CCVPE is independent of the number of position samples, its performance is shown as a horizontal line.
    }
    \label{fig:figureD}
\end{figure}

\subsection{Effect of candidate pose sampling density}
An ablation study is conducted to examine the impact of candidate pose sampling density, with the analysis separated into position and orientation sampling. For position sampling, three spatial resolutions are compared: $10{\times}10$, $15{\times}15$, and $20{\times}20$. For orientation sampling, four candidate counts are evaluated: 20, 40, 64, and 70. These sampling configurations are selected with reference to prior works~\cite{lentsch2023slicematch, xia2023convolutional}.

\cref{fig:figureD} shows that increasing the number of candidate poses leads to reduced median errors in both position and orientation. This trend highlights the benefit of finer sampling resolution. Although denser sampling improves accuracy, it also increases computational cost. \cref{sec:tradeoff_sampling} examines this trade-off in more detail. We also compare two global descriptor-based approaches: SliceMatch~\cite{lentsch2023slicematch} and CCVPE~\cite{xia2023convolutional}. The proposed method consistently improves over all baselines across the tested sampling densities, demonstrating the superiority of our approach.

\begin{figure}[t]
    \centering
    \includegraphics[width=\columnwidth]{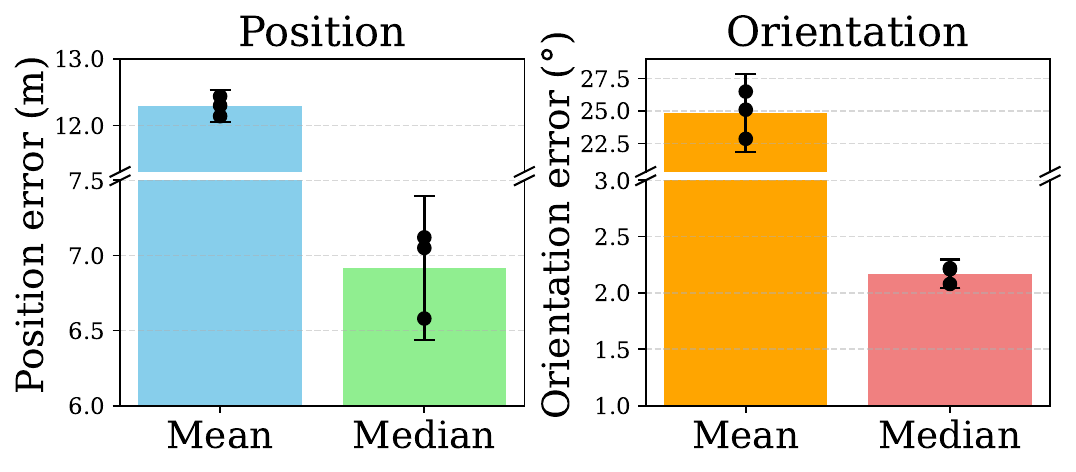}
    \caption{Error statistics across five random seeds for the proposed method. Error bars indicate $\pm2\sigma$ standard deviations, computed from five independent runs.
    }
    \label{fig:seed_error_bar_summary}
\end{figure}

\begin{table*}[t]
    \caption{Inference memory usage and runtime of the proposed method and prior approaches on the KITTI~\cite{geiger2013vision} dataset, including the impact of different sampling densities. Sampling density is defined as the number of candidate poses, computed as the product of position and orientation sampling sizes. All results are measured using the official implementations on an RTX A5000 GPU with 24\,GB memory.}
    \label{tab:computational}
  \centering
  \footnotesize
  \begin{threeparttable}
    \vspace{-2mm}
    \setlength{\tabcolsep}{4pt}
    \begin{adjustbox}{width=\linewidth}
        \begin{tabular}{lcccccc}
            \toprule
            Method & Backbone & Sampling density & Med. position error & Med. orientation error
            & GPU memory usage (GB) & Inference runtime (FPS) \\
            \midrule
            HighlyAccurate~\cite{shi2022beyond} & VGG16 & - & 16.02 & 89.85
            & 7.1 & 3 \\
            CCVPE~\cite{xia2023convolutional} & EfficientNet-B0 & - & 10.98 & 63.84
            & 5.2 & 20 \\
            DenseFlow~\cite{song2023learning} & ResNet18 & - & 18.84 & 42.04
            & 6.4 & 9 \\
            FG2~\cite{xia2025fg} & DINOv2 & - & 11.72 & 90.42
            & 3.5 & 6 \\
            \midrule
            VIRD (Ours) & VGG16 & 10$\times$10$\times$70 & 9.40 & 2.63
            & 4.8 & 27 \\
            VIRD (Ours) & VGG16 & 15$\times$15$\times$70 & 7.50 & 2.35
            & 6.6 & 18 \\
            VIRD (Ours) & VGG16 & 20$\times$20$\times$70 & 7.05 & 2.22
            & 9.0 & 12 \\
            VIRD (Ours) & EfficientNet-B0 & 20$\times$20$\times$70 & 5.41 & 1.87 
            & 9.0 & 13 \\
            \bottomrule
        \end{tabular}
    \end{adjustbox}
  \end{threeparttable}
\end{table*}

\section{Error Statistics across Random Seeds}
\label{sec:suppl_error_bar}
To assess the robustness of the proposed method, performance variation is reported across the five independent runs using different random seeds. \cref{fig:seed_error_bar_summary} visualizes the mean and median errors for both position and orientation, providing a statistical perspective on the results presented in \cref{tab:kitti}. Error bars represent $\pm2\sigma$ standard deviations, computed as the sample standard deviation across five independent runs. This captures variability due to random initialization and training dynamics. While the normality assumption is not formally tested due to the limited number of samples, the visualization provides a practical view of the method’s stability across repeated runs.

\section{Computational Resources and Runtime}
\label{sec:suppl_compute}
\subsection{Training and inference}
We report the computational cost during both training and inference. All experiments were conducted on a local workstation equipped with an RTX A5000 GPU with 24\,GB of memory. A batch size of 4 was used during training and 1 during inference. During training, the peak GPU memory usage reached approximately 24\,GB on both the KITTI and VIGOR datasets. Using the VGG16 backbone, training required about 15 hours on KITTI and 38 hours on VIGOR.

\cref{tab:computational} summarizes GPU memory usage and runtime during inference on the KITTI dataset. With the VGG16 backbone and a sampling density of $20\times20\times70$, VIRD requires 9.0\,GB of GPU memory and achieves 12 frames per second (FPS). Using EfficientNet-B0 yields similar memory usage while slightly improving runtime to 13 FPS. Compared with prior methods, the proposed approach offers substantially higher accuracy while maintaining comparable memory usage and fast inference speed. These results demonstrate that VIRD is computationally efficient and practical for real-world deployment.

\subsection{Trade-off between sampling density and efficiency}
\label{sec:tradeoff_sampling}
\cref{tab:computational} also illustrates how increasing the candidate pose sampling density affects computational cost. Higher sampling densities lead to increased memory usage and longer inference times, reflecting a clear trade-off between accuracy and efficiency. Therefore, users may refer to \cref{tab:computational} to select a sampling configuration that best meets their performance and runtime constraints.

\section{Extra Qualitative Results}
\subsection{Attention weights visualization}
\begin{figure*}[ht]
    \begin{subfigure}{\textwidth}
        \centering
        \includegraphics[width=\textwidth]{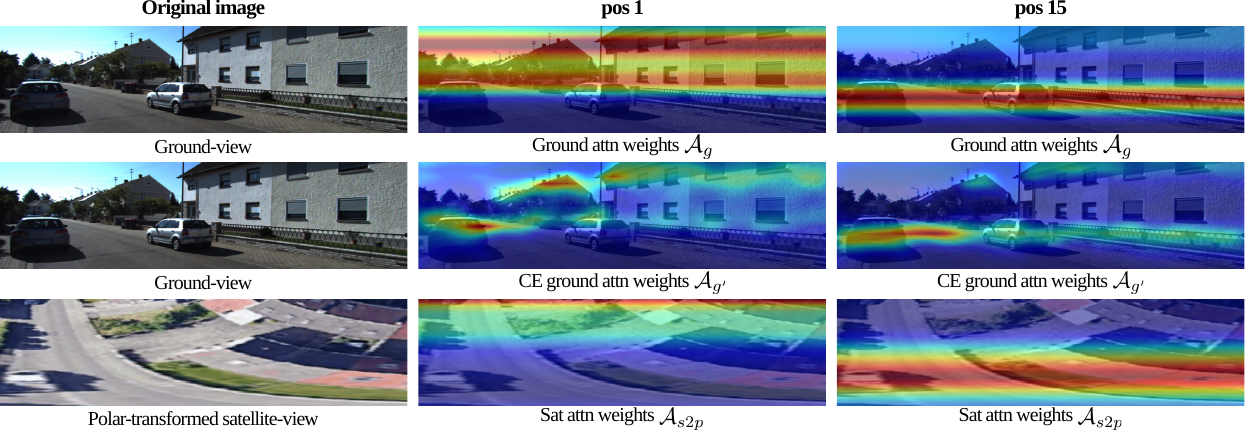}
        \caption{Example 1}
        \label{fig:figureF_a}
    \end{subfigure}
    \par\medskip
    \begin{subfigure}{\textwidth}
        \centering
        \includegraphics[width=\textwidth]{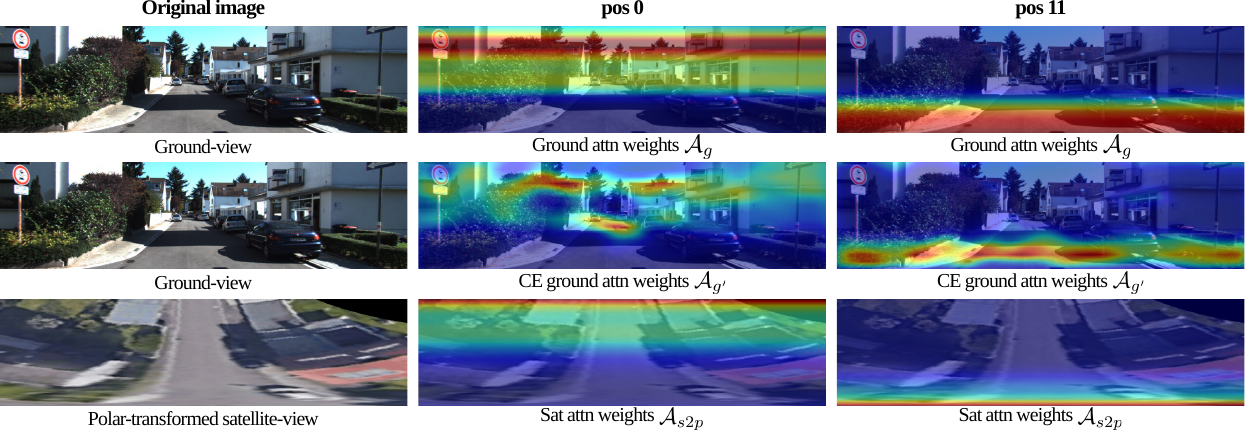}
        \caption{Example 2}
        \label{fig:figureF_b}
    \end{subfigure}
    \par\medskip
    \begin{subfigure}{\textwidth}
        \centering
        \includegraphics[width=\textwidth]{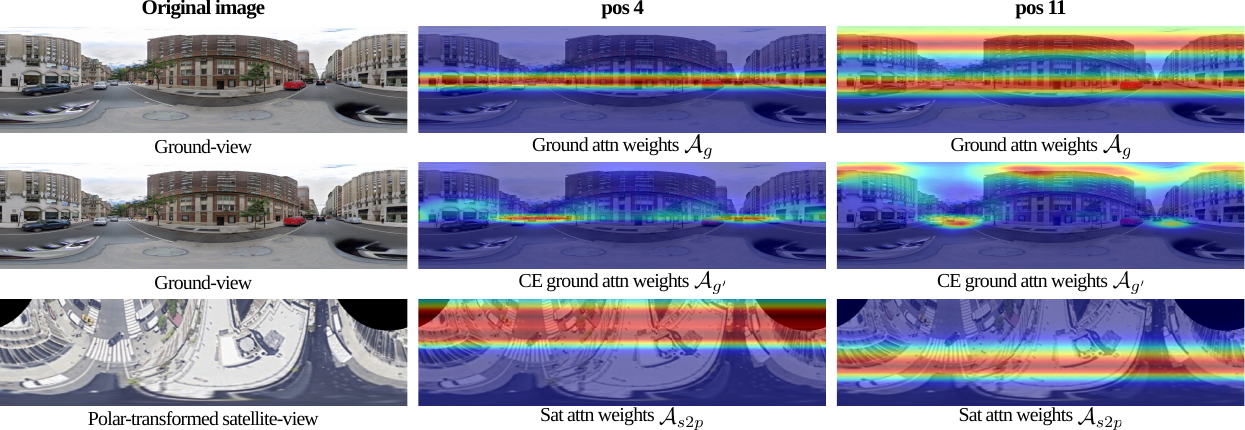}
        \caption{Example 3}
        \label{fig:figureF_c}
    \end{subfigure}
    \caption{Visualization of the attention weights $\AttnW{g}$, $\AttnW{g'}$, and $\AttnW{s2p}$ at selected positions in the shared virtual positional encoding space. CE and pos denote the context-enhanced attention and the positions in the shared virtual positional encoding space, respectively. Regions with stronger activations are highlighted in red.}
    \label{fig:figureF}
\end{figure*}

We provide additional qualitative examples of attention weights.
\cref{fig:figureF} visualizes $\AttnW{g}$, $\AttnW{g'}$, and $\AttnW{s2p}$, which are computed by applying context-enhanced positional attention to ground and polar-transformed satellite features. The positional attention weights $\AttnW{g}$ and $\AttnW{s2p}$, which are computed without context enhancement, exhibit consistent patterns along the horizontal direction due to learning a shared vertical transformation. In contrast, the context-enhanced attention $\AttnW{g'}$ adaptively incorporates contextual information for each horizontal position, resulting in more diverse horizontal attention distributions. Overall, these visualizations indicate that the proposed approach effectively captures meaningful spatial correspondence between ground and satellite views.

\subsection{Reconstructed images visualization}
\label{sec:qualitative_recon}
\begin{figure*}[ht]
    \begin{subfigure}{\textwidth}
        \centering
        \includegraphics[width=\textwidth]{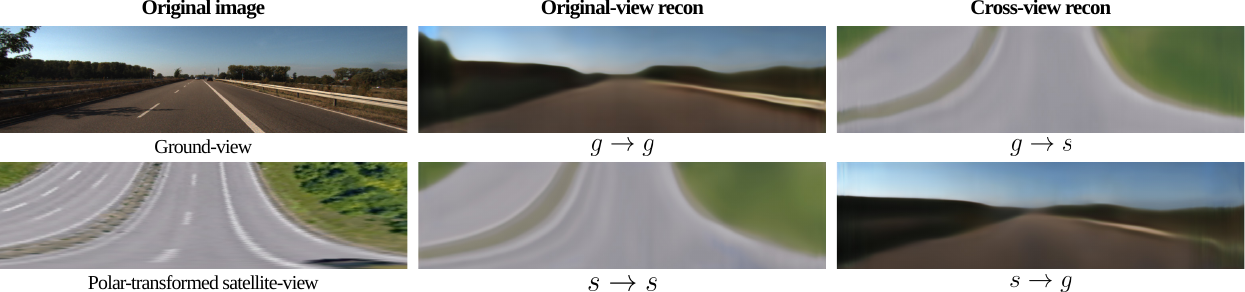}
        \caption{Example 1}
        \label{fig:figureG_a}
    \end{subfigure}
    \par\medskip
    \begin{subfigure}{\textwidth}
        \centering
        \includegraphics[width=\textwidth]{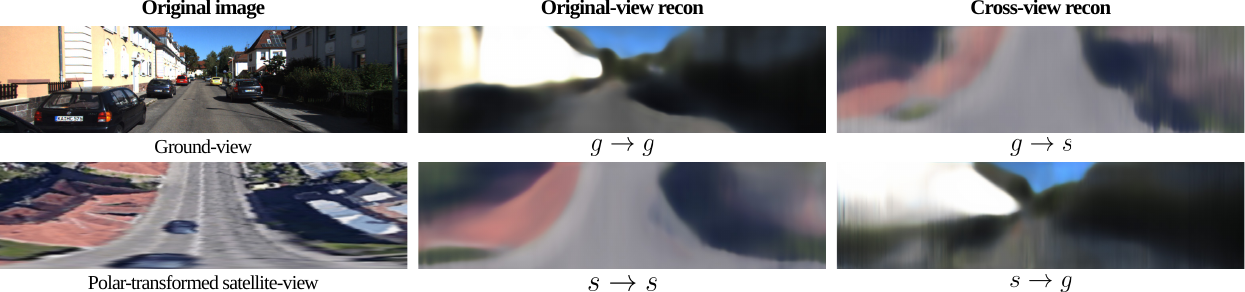}
        \caption{Example 2}
        \label{fig:figureG_b}
    \end{subfigure}
    \par\medskip
    \begin{subfigure}{\textwidth}
        \centering
        \includegraphics[width=\textwidth]{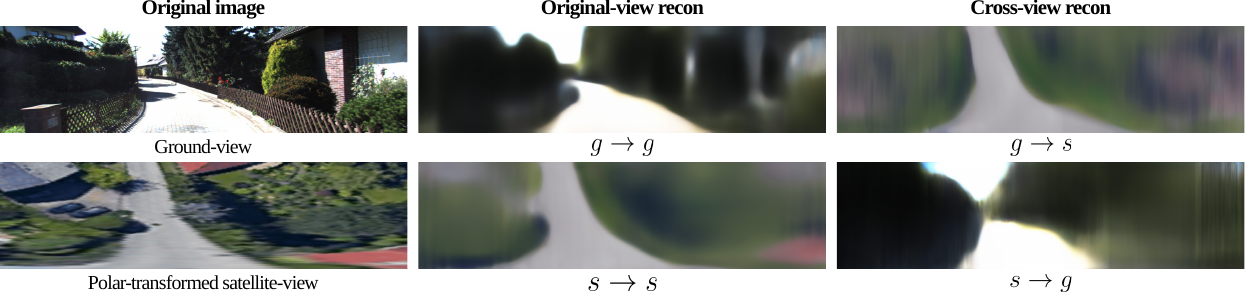}
        \caption{Example 3}
        \label{fig:figureG_c}
    \end{subfigure}
    \caption{Visualization of reconstructed images from ground and satellite descriptors in both original and cross views. Each subfigure is denoted as $i \to j$, where $i, j \in \{s, g\}$ indicate the source and target views, respectively. The pairs $g \to g$ and $s \to s$ indicate original-view reconstructions, whereas $g \to s$ and $s \to g$ represent cross-view reconstructions.}
    \label{fig:figureG}
\end{figure*}

We present additional qualitative examples of reconstructed images.
\cref{fig:figureG} visualizes the reconstruction results of the ground and satellite descriptors in both original and cross views. The reconstructed outputs recover road layouts and capture coarse building structures, while naturally suppressing view-specific elements such as vehicles and side roads. These observations indicate that the view-reconstruction loss effectively drives the descriptors toward view-invariant representations that are consistently preserved across ground and satellite views.

\section{Societal Impact}
\label{sec:suppl_social_impact}
The proposed method advances cross-view localization by enabling accurate ground-to-satellite matching in GNSS-denied environments, such as dense urban areas, where satellite imagery is available. This functionality supports critical applications, including autonomous navigation and disaster response. However, it raises potential privacy concerns and risks of unintended uses, such as surveillance or location-based profiling. These concerns highlight the importance of considering ethical deployment practices and conducting privacy assessments in downstream applications.

\section{Limitations}
\label{sec:suppl_limit}

A key limitation of the proposed method is its dependence on the number of candidate poses sampled during inference. Since both model accuracy and computational efficiency are sensitive to the sampling resolution, the method requires manual tuning of the sampling density to meet application-specific constraints, which limits its scalability in practice. In addition, similar to most existing cross-view pose estimation approaches, our method assumes that roll and pitch angles can be neglected. Although this assumption holds in relatively flat urban environments, it may lead to significant errors in challenging terrains such as mountainous regions or areas with strong elevation variations. To address these limitations, future work could focus on reducing the method’s reliance on candidate pose sampling resolution and extending the approach toward estimating the full 6-DoF pose without assuming fixed roll and pitch.


\end{document}